\documentclass[10pt,twocolumn,letterpaper]{article}

\usepackage[pagenumbers]{cvpr} 

\usepackage{bm}
\usepackage[ruled,vlined,linesnumbered]{algorithm2e}

\usepackage{url}
\usepackage{algorithmic}
\usepackage{graphicx,wrapfig,lipsum}
\usepackage{booktabs}
\usepackage{colortbl}  
\usepackage{xcolor}
\usepackage{multirow} 
\usepackage{pifont}
\usepackage{amsmath}
\usepackage{amsfonts}
\usepackage{amssymb}
\usepackage{float}
\usepackage{siunitx}
\usepackage{pifont}
\usepackage[font=normalsize,labelfont=bf,tableposition=top]{caption}
\usepackage{csquotes} 
\definecolor{aliceblue}{rgb}{0.94, 0.97, 1.0}
\definecolor{deeppink}{RGB}{255,20,147}
\definecolor{mygray}{gray}{.9}
\definecolor{mygray2}{gray}{.6}
\usepackage[table]{xcolor}

\newcommand{\thickhline}{%
    \noalign {\ifnum 0=`}\fi \hrule height 0.8pt
    \futurelet \reserved@a \@xhline
}

\usepackage{amsmath,amsfonts,bm}

\def\eqref#1{equation~\ref{#1}}

\def\1{\bm{1}}

\def\rvz{{\mathbf{z}}}

\DeclareMathAlphabet{\mathsfit}{\encodingdefault}{\sfdefault}{m}{sl}
\SetMathAlphabet{\mathsfit}{bold}{\encodingdefault}{\sfdefault}{bx}{n}

\newcommand{\myparagraph}[1]{\noindent\textbf{#1}\,\,}

\newcommand{\sourceInput}{\mathbf{s}}
\newcommand{\reasonInput}{\mathbf{r}}
\newcommand{\targetOutput}{\mathbf{e}}

\newcommand{\vaeEncoder}{\mathcal{E}}
\newcommand{\vaeDecoder}{\mathcal{D}}

\newcommand{\OursMethod}{VideoCoF}
\newcommand{\OursBench}{VideoCoF-Bench}

\definecolor{oursrow}{RGB}{242,247,242} 

\definecolor{cvprblue}{rgb}{0.21,0.49,0.74}
\usepackage[pagebackref,breaklinks,colorlinks,citecolor=cvprblue]{hyperref}
\usepackage{marvosym}
\def\paperID{957} 
\def\confName{CVPR}
\def\confYear{2026}

\title{VideoCoF: Unified Video Editing with Temporal Reasoner}

\author{
    Xiangpeng Yang\textsuperscript{1} \quad
    Ji Xie\textsuperscript{2} \quad
    Yiyuan Yang\textsuperscript{1} \quad
    Yue Ma\textsuperscript{3} \quad
    Yan Huang\textsuperscript{1\textbf{\Letter}} \quad
    Min Xu\textsuperscript{1} \quad
    Qiang Wu\textsuperscript{1} \\
    \textsuperscript{1}University of Technology Sydney \qquad
    \textsuperscript{2}Zhejiang University \qquad
    \textsuperscript{3}HKUST \\
    {
        \centerline{\href{https://videocof.github.io/}{\texttt{https://videocof.github.io/}}}
    }
}

\begin{document}

\twocolumn[{
\begin{center}
\maketitle
\vspace{-2em}
    \captionsetup{type=figure}
    \includegraphics[width=0.95\textwidth]{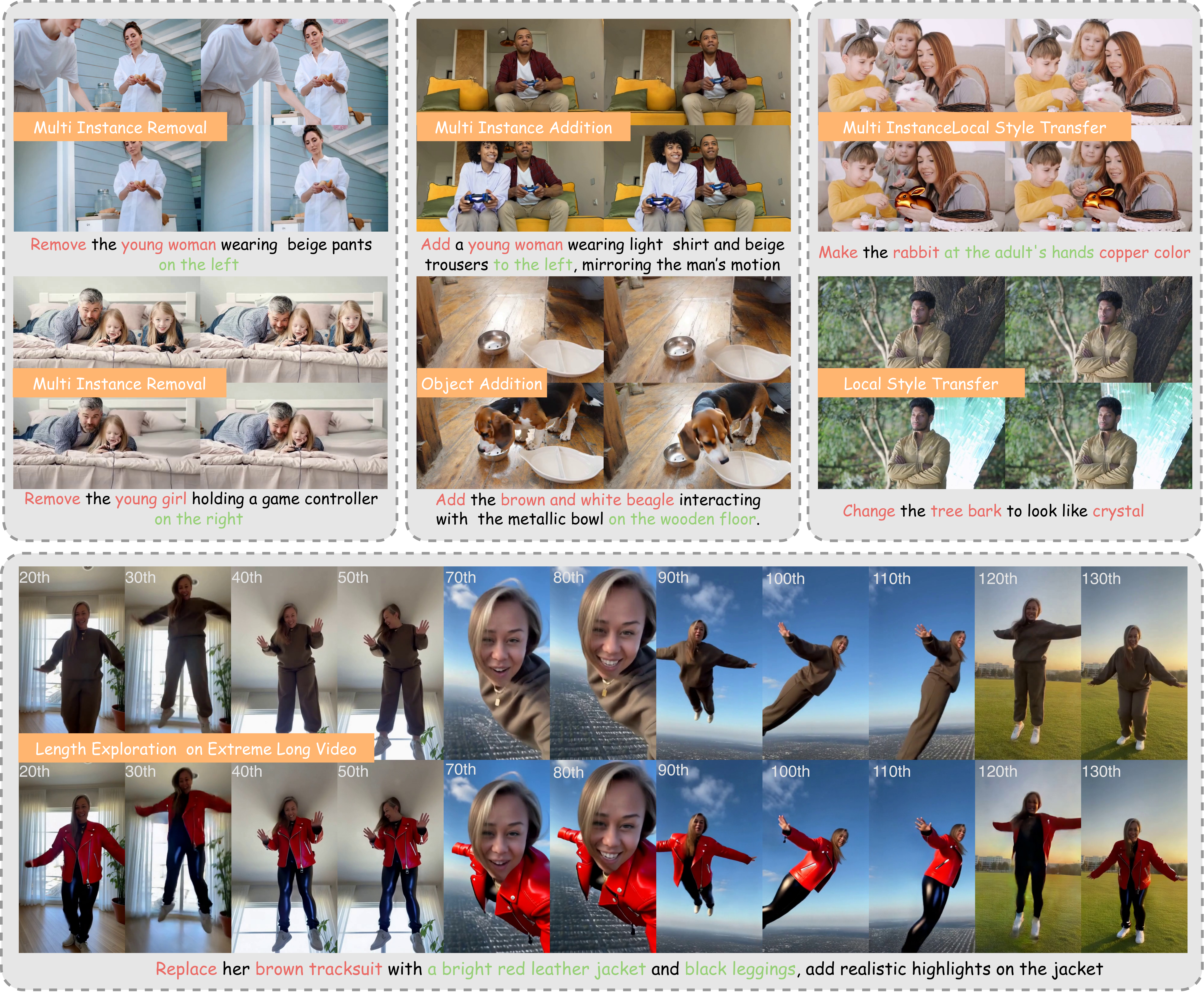}
    \vspace{-0.5em}
    \caption{\OursMethod's video editing capabilities emerge from its \textbf{seeing, reasoning, then editing framework}. Trained on only \textbf{50k} data (33 frames), this teaser shows multi-instance editing and robust $4\times$ length generalization.}
    \label{fig:teaser}
    \vspace{-0.5em}
\end{center}
}]

\begingroup
\renewcommand{\thefootnote}{}
\footnotetext{\Letter~Corresponding author.}
\endgroup

\begin{abstract}

\vspace{-1em}
Existing video editing methods face a critical trade-off: expert models offer precision but rely on task-specific priors like masks, hindering unification; conversely, unified temporal in-context learning models are mask-free but lack explicit spatial cues, leading to weak instruction-to-region mapping and imprecise localization. To resolve this conflict, we propose \textbf{VideoCoF}, a novel \textbf{Chain-of-Frames} approach inspired by Chain-of-Thought reasoning. 
VideoCoF enforces a ``see $\rightarrow$ reason $\rightarrow$ edit" procedure by compelling the video diffusion model to first predict \textbf{reasoning tokens} (edit-region latents) before generating the target video tokens. This explicit reasoning step removes the need for user-provided masks while achieving precise instruction-to-region alignment and fine-grained video editing. Furthermore, we introduce a RoPE alignment strategy that leverages these reasoning tokens to ensure motion alignment and enable length extrapolation beyond the training duration. We demonstrate that with a minimal data cost of only 50k video pairs, VideoCoF achieves state-of-the-art performance on VideoCoF-Bench, validating the efficiency and effectiveness of our approach. Our code, weight, data are available at \href{https://github.com/knightyxp/VideoCoF}{https://github.com/knightyxp/VideoCoF}.

\end{abstract}
    
\section{Introduction}
\label{sec:intro}

The development of Video Diffusion Models (VDM) ~\citep{tuneavideo, cogvideox, hunyuanvideo, wanx} has enabled high-fidelity video generation across a wide range of concepts. Building on these advances, video editing methods support users in designing video by adding \citep{tu2025videoanydoor}, removing \citep{zi2025minimax, diffueraser}, swapping \citep{gu2024videoswap,yang2025videograin} visual concepts, and performing global style transformation \citep{stylemaster}.

Current video editing methods mainly follow two strategies: (i) \textbf{expert models} ~\citep{controlnet,videopainter,diffueraser,tu2025videoanydoor,yang2025videograin}, which use adapter-based modules to feed \emph{external masks} into the video generation model, yielding precise, localized edits but requiring additional inputs and per-task overhead; and (ii) \textbf{unified temporal in-context learning models} ~\citep{icve,unic,editverse}, which concatenate source tokens with noised edit tokens along the temporal dimension and use self-attention mechanism to guide the edit. However, without explicit spatial cues, these models often exhibit weak accuracy, especially in cases that need multi-instance recognition or spatial reasoning (Fig.~\ref{fig:motivation}, left). In short, there is a \emph{trade-off}: expert models are accurate but mask-dependent, while unified in-context models are mask-free but less precise; This raises a critical question: \textbf{Can we maintain former's precision and latter's unification without the mask dependency?}


Inspired by Chain-of-Thought (CoT) multi-step reasoning~\citep{wei2022chain}, we \emph{compel} the video diffusion model to first predict the edit region and then perform the edit, enforcing a ``\textbf{see}~$\rightarrow$~\textbf{reason}~$\rightarrow$~\textbf{edit}'' procedure. Accordingly, we propose \textbf{\emph{\OursMethod}}, a Chain-of-Frames approach that predicts \emph{reasoning tokens} (edit-region latents) before generating the target video tokens, thereby removing the need for user-provided masks while achieving precise instruction-to-region alignment.
To explicitly model the reasoning process, we leverage visual grounding, which is naturally suited to simulating reasoning about the edit region.
Empirically, we find a soft, gradually highlighted grayscale region is the most effective reasoning format.
Additionally, we introduce a RoPE alignment strategy. 
By explicitly accounting for the reasoning latent, we reset the temporal indices of the edited video’s rotary position embeddings to match those of the source video, ensuring motion alignment and length extrapolation.

To holistically evaluate fine-grained video editing, we further construct VideoCoF-Bench. \OursMethod~ trained on only \textbf{50k} video pairs, outperforms a strong baseline ICVE \citep{icve} that uses \(\sim\)1M pretraining videos plus 150k for finetuning. Specifically, we improve the instruction-following score by \textbf{+15.14\%} and the success ratio by \textbf{+18.6\%}. Our contributions can be summarized as follows: 

\begin{figure}[t]
\centering
\scriptsize
\includegraphics[width=\linewidth]{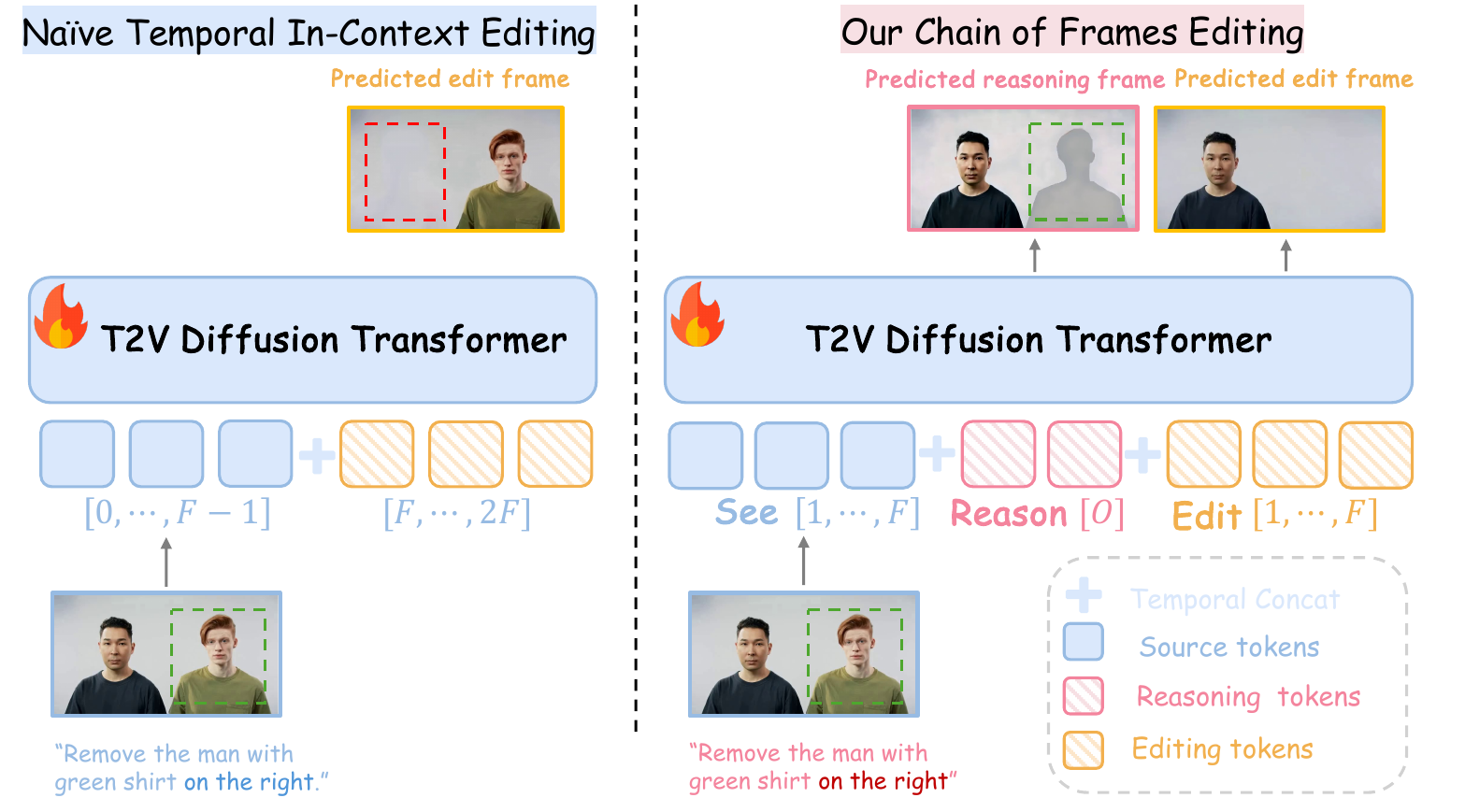}  
 \vspace{-1.0 em}
\scriptsize
\caption{Illustration of the difference between previous methods and our \OursMethod. We enhance the editing accuracy by forcing the video diffusion model to first predict the editing area, and then perform the editing.}
\vspace{-1.0 em}
\label{fig:motivation}
\end{figure} 

\begin{itemize} 
\item We propose VideoCoF, the first framework to introduce a Chain of Frames approach to video editing, enabling temporal reasoning for fine-grained video editing.
\item Building on \OursMethod, we explore an effective reasoning format for video diffusion models, and introduce a RoPE alignment strategy that allows generalization to longer frames beyond the training duration.
\item We demonstrate that with a minimal data cost (only \textbf{50k} video pairs), we achieve state-of-the-art quantitative and qualitative performance on \OursBench, validating the efficiency and effectiveness of our approach. 
\end{itemize}

\begin{figure*}[t!]
    \vspace{-2mm}
    \centering
    \includegraphics[width=0.99\linewidth]{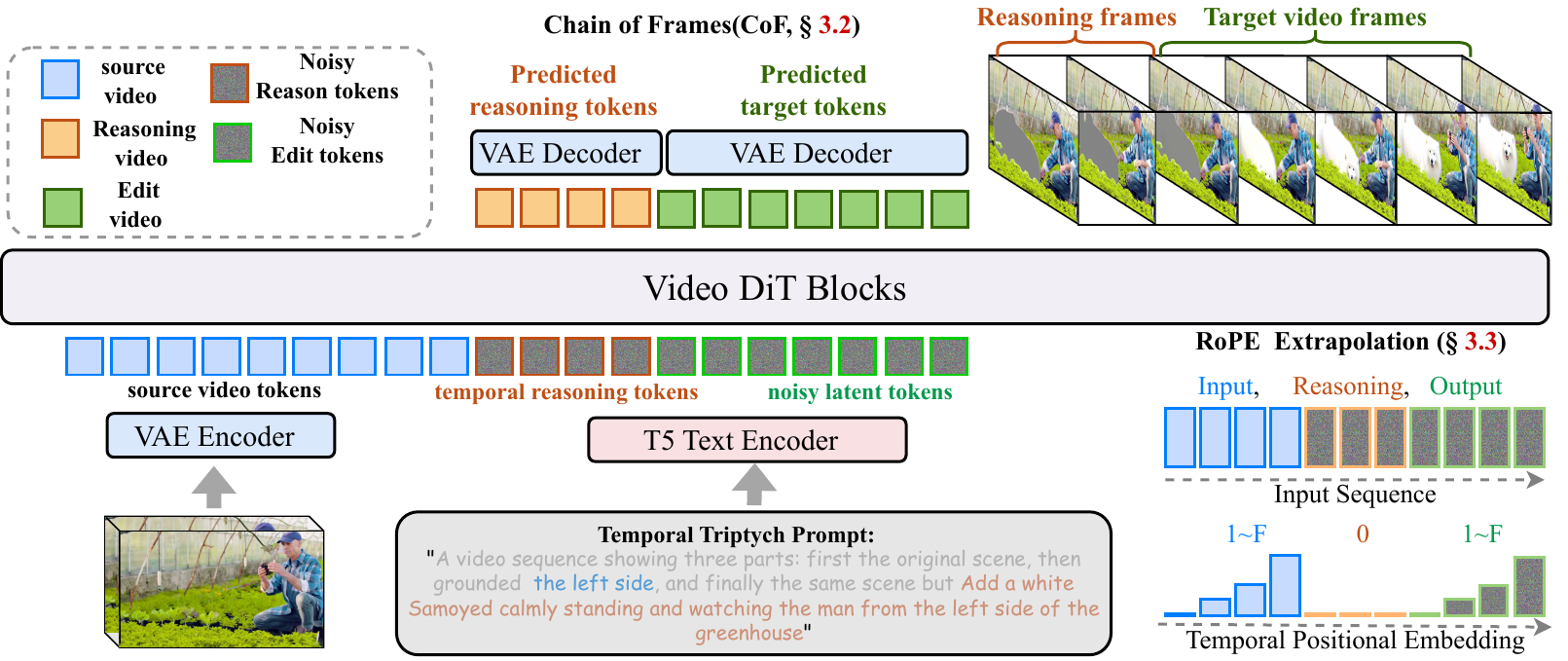}
    \vspace{-2mm}
    \caption{\textbf{Overview of \OursMethod~ framework.}
    Our model processes source (blue), reasoning (orange), and target (green) tokens in a unified sequence to ``reason" then ``edit". 
    \textbf{Bottom right:} Our RoPE design enables length extrapolation.
    }
    \vspace{-3mm}
\label{fig:framework}
\end{figure*}

\section{Related Work}
\label{sec:related work}

\noindent \textbf{Video Editing Methods.}
Early training-free video editing methods \citep{tuneavideo,fatezero,geyer2023tokenflow,yang2024eva,chen2025contextflow,long2025follow} rely on inversion and consistency techniques, but often lack precise control and struggle with complex edits. Recent training-based methods \citep{instructpix2pix, insv2v, cai2025omnivcus, liu2025generative, ma2025followyourmotion, shen2025follow} have become the dominant paradigm, offering higher quality and edit diversity. Meanwhile, concurrent works \citep{yu2025veggie,univideo,instructx,liu2025tuna,lin2025exploring} integrate MLLMs to guide the editing process, though this adds significant training and inference cost, which our pure VDM approach avoids.

\noindent \textbf{In-Context Video Editing.}
Recently, in-context learning (ICL) has emerged as a promising paradigm for unified editing \citep{zhang2025context,fulldit,omnigen}. Methods like UNIC \cite{unic} and ICVE \citep{icve} concatenate video conditions along the temporal axis to perform ICL. However, these methods are often limited by mask requirements \citep{unic} or, as we identify, suffer from fundamental issues with editing accuracy and a lack of length extrapolation due to their naive temporal concatenation. While EditVerse \citep{editverse} also explored unified in-context learning, it was built on a LLaMA-style DiT backbone, whereas our work explores these capabilities within a standard video diffusion transformer.

\noindent \textbf{Chain of Thought in Vision.}
Chain-of-Thought (CoT) prompting \citep{wei2022chain, kojima2022large} elicits multi-step reasoning in LLMs by having them ``think step-by-step." This concept of emergent reasoning has also been identified in large video generative models \citep{wiedemer2025video, chen2025univid} that can solve visual puzzles. However, how to leverage visual reasoning for the task of unified video editing remains unexplored. In this work, we investigate whether generative video models can perform a ``chain of frames" reasoning to achieve this.
\section{Methods}
\label{sec:method}



\subsection{VideoCoF Framework}
\label{sec:framework}
As illustrated in Figure \ref{fig:framework}, VideoCoF employs a VideoDiT \cite{wanx} for unified video editing. 
We model editing as a reasoning-then-generation process: the model first reasons where to edit, then generates the intended content in that area. We call this process \textbf{``Chain of Frames (CoF)''} (Sec \ref{sec:CoF}).
All visual inputs (source, reasoning, and target frames) are encoded separately by a Video VAE and then concatenated temporally. The unified frame sequence is then fed into the model, performing unified in-context learning via self-attention and language control via cross-attention.
To enable video alignment and variable-length inference, we revisit the design of positional encoding. We adapt the temporal RoPE for source-to-target alignment and reasoning tokens' RoPE for explicit spatial guidance (Sec \ref{sec:rope}). Subsequent sections detail the training and inference paradigm (Sec \ref{sec:train and infer}), and the data curation pipeline (Sec \ref{sec:data}).


\subsection{Chain of Frames}
\label{sec:CoF}

\myparagraph{Seeing, Reasoning, then Editing.}
Previous video in-context editing methods, such as UNIC \citep{unic}, ICVE \citep{icve}, or EditVerse \cite{editverse}, perform in-context learning by temporally concatenating clean source video tokens with noised editing video tokens. However, this approach lacks an explicit constraint mapping the editing instruction to the specific editing region, leading to editing accuracy problems, as shown in Fig~\ref{fig:motivation}. 
Recently, VDM have been shown to possess reasoning capabilities, as demonstrated in \citep{wiedemer2025video}. 
Inspired by this, we explicitly model the reasoning tokens, forcing the model to actively learn the relationship between the editing instruction and the target edit region first. 
The edit is then executed \emph{after} reasoning, following a ``seeing, reasoning, then editing” process.

Inspired by Chain of Thought prompting in Large Language Models (LLMs) \citep{wei2022chain}, we argue that a video generative model should also have an analogous chain-reasoning ability. 
Given the generative priors in video editing, the visual-chain should be progressive, moving from the original video to a visual reference of the editing region, and finally to the edited video.
Visual grounding is naturally suitable for this representation.
Since video diffusion models are often insensitive to grounding masks (e.g., black or white pixels), we instead use a gray highlight to delineate the ``grounding region," which is also evidenced in \cite{offsetNoise}.
Finally, the gray-highlighted area is used as the ground truth for the reasoning frames, teaching the diffusion model to reason about where the edit should occur.





Consequently, the entire video editing task is reformulated as a chained process: first ``seeing'' the original video, then ``reasoning'' by predicting the grounding region, and finally ``editing'' to generate the new video content within that specified area. We call this \textbf{Chain of Frames (\textit{CoF})}.

Let $\vaeEncoder(\cdot)$ denote the video VAE encoder. We use $F$ and $L$ for frames in the source/target and reasoning latent space, respectively, and denote channel, height, and width by $C$, $H$, and $W$.
Given a triplet source-reasoning-target video pair $\{\sourceInput,\reasonInput, \targetOutput\}$, we first encode them into latent representations. The source $\sourceInput$ and target video $\targetOutput$ yield latent ${z_{s}} = \vaeEncoder(\sourceInput)$ and ${z_{e}} = \vaeEncoder(\targetOutput)$, both with shape $\mathbb{R}^{F \times C \times H \times W}$. The reasoning video $\reasonInput$ yields a latent ${z_{r}} = \vaeEncoder(\reasonInput)$ with shape $\mathbb{R}^{L \times C \times H \times W}$. This separate encoding ensures intra-causal relations and inter-video independence. Then, we perform temporal concatenation to get the unified representation:
\begin{equation}
    \mathbf{z}_{full}^{(t)} = 
\underbrace{z_{s}^{(0)}}_{\text{seeing}}
\;\Vert\;
\underbrace{z_{r}^{(t)}}_{\text{reasoning}}
\;\Vert\;
\underbrace{z_{e}^{(t)}}_{\text{editing}}
\quad\in\mathbb{R}^{(F + L + F)\times C\times H\times W},
\end{equation}
where the ${z_s={\mathbf{z}}_{0:F-1}^{(0)}}$ denotes anchoring the source video latent at timestep 0.
${z_{r}={\mathbf{z}}_{F:F+L-1}^{(t)}}$ and  ${z_{e}={\mathbf{z}}_{F+L:2F+L-1}^{(t)}}$ mean the reasoning and target noised video latents at timestep t.
At each denoising step, only the $L+F$ reasoning and target frames are denoised, and the source video latents are kept clean.

\subsection{RoPE Design for Length Extrapolation}
\label{sec:rope}

\begin{figure}
    \centering
    \includegraphics[width=\linewidth]{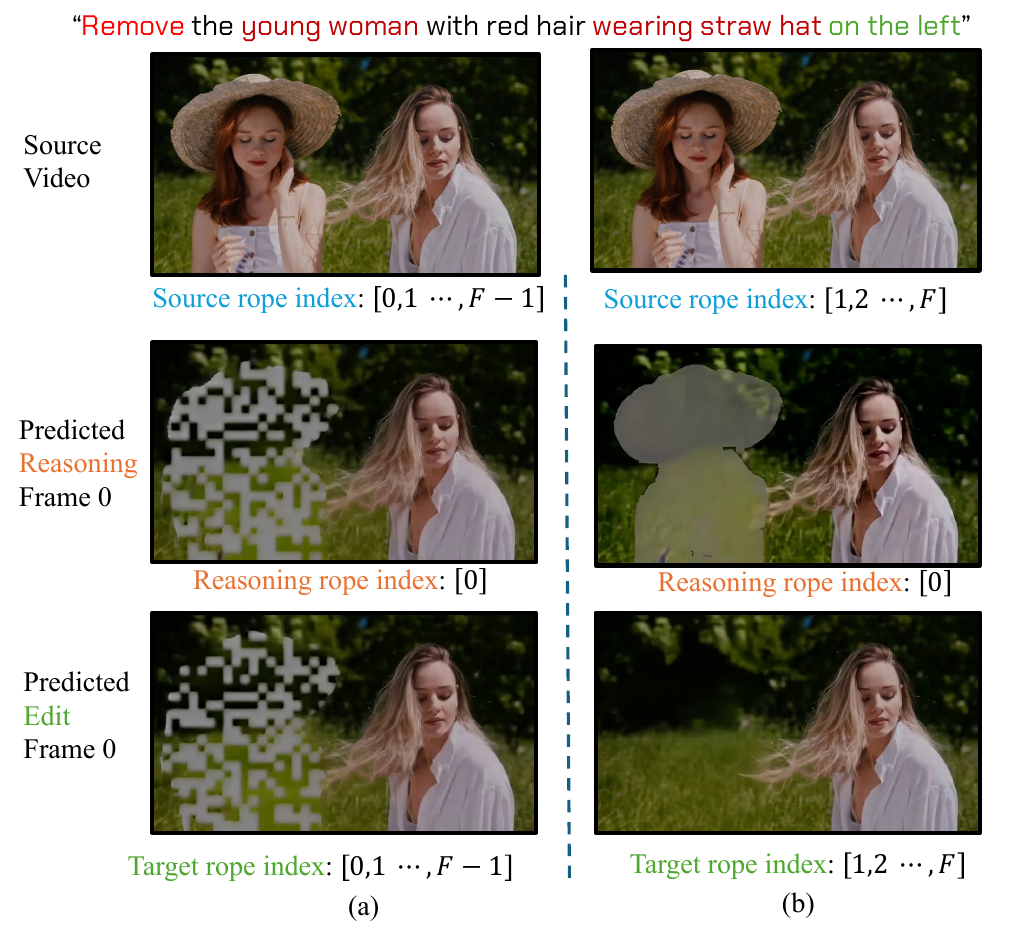}
    \vspace{-1em}
    \caption{How our RoPE design avoids index collision.\vspace{-15pt}}
    \label{fig:reason_vis}
\end{figure}

In VideoDiT, 3D factorized RoPE \cite{rope} provides spatio-temporal positions. A naive in-context learning approach applies sequential temporal indices (e.g., $0$ to $2F-1$) across concatenated source and target videos. However, this hinders video length extrapolation, as the model overfits to a static $[0, F-1] \to [F, 2F-1]$ mapping and fails to generalize to videos longer than $F$ frames.

A better strategy is to repeat the temporal indices. For our CoF triplet (consider $L=1$ for reasoning frame), a straightforward reset configuration is to assign temporal indices: $[0, F-1]$ to the source, ``0'' to the reasoning frame, and $[0, F-1]$ to the target.

However, as illustrated in Figure~\ref{fig:reason_vis} (a), this naive reset leads to index collisions at temporal position 0, shared by the source, reasoning, and target frames. This overlap introduces visual artifacts that propagate from the reasoning tokens into the first target frame.

To resolve this index collision, we set the temporal indices for both the source video and the target video to the range $\mathbf{[1, F]}$, while keeping the reasoning frame's temporal index at $\mathbf{0}$. This isolates the reasoning token and prevents artifact leakage while maintaining length generalization.

\subsection{Training and Inference Paradigm}
\label{sec:train and infer}

\begin{figure*}[t]
    \centering
    \includegraphics[width=\linewidth]{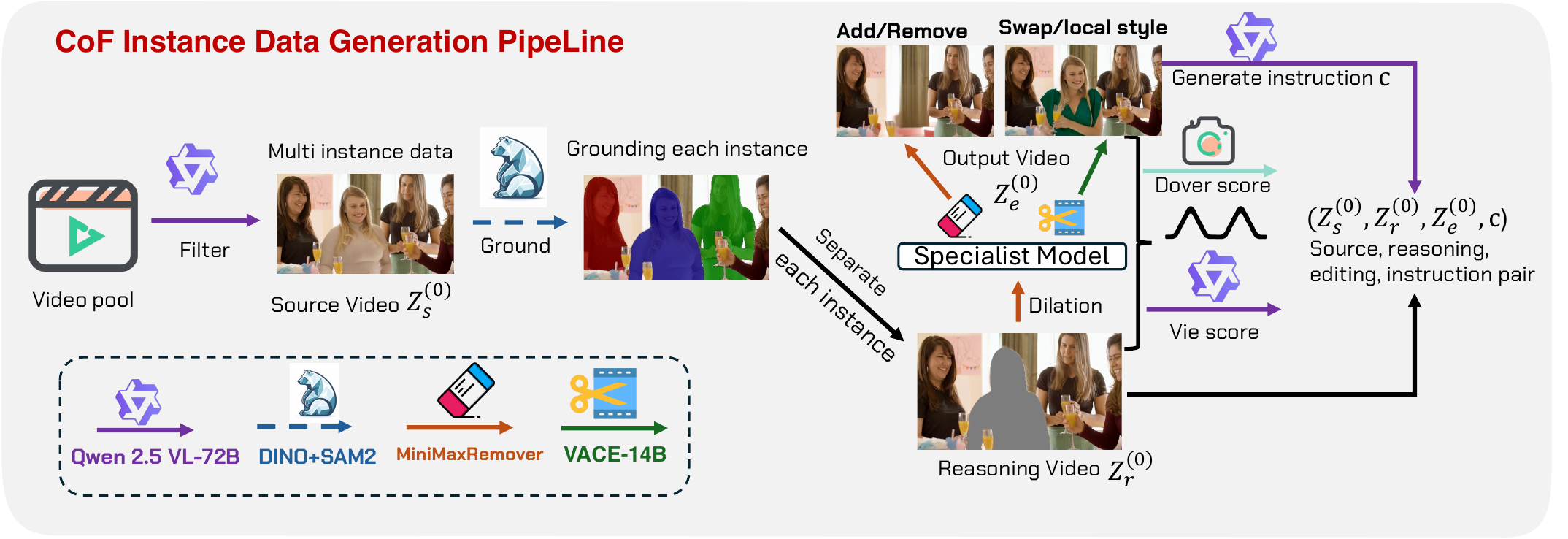}
    \vspace{-25pt}
    \caption{Our data curation pipeline for multi-instance data. \vspace{-15pt}}
    \label{fig:data_pipe}
\end{figure*}

\begin{algorithm}[h] 
\LinesNotNumbered
\LinesNotNumbered
\caption{Chain of Frame (CoF) Training}
\label{alg:train}
\small
\SetKwInput{KwIn}{Input}\SetKwInput{KwOut}{Output}
\KwIn{Dataset $\mathcal{D}$ with tuples $(\mathbf{z_{s}}^{(0)},\mathbf{z_{r}}^{(0)},{\mathbf{z_{e}}}^{(0)},\mathbf{c})$}
\KwOut{Fine-tuned parameters $\bm{\theta}$}
\BlankLine
\ForEach{\textnormal{minibatch } $(\mathbf{z_{s}}^{(0)},\mathbf{z_{r}}^{(0)},{\mathbf{z_{e}}}^{(0)},\mathbf{c})\sim\mathcal{D}$}{
    \ForEach{\textnormal{sample in minibatch}}{
        $\rvz_{full}^{(0)} \leftarrow \mathbf{z_{s}}^{(0)} \Vert \mathbf{z_{r}}^{(0)} \Vert \mathbf{z_{e}}^{(0)}$\;
        Sample $t \sim \mathcal{U}[0, 1]$\;
        
        Sample $\boldsymbol{\varepsilon} \sim \mathcal{N}(\mathbf{0},\mathbf{I})$ with the same shape $\rvz_{full}^{(0)}$\;
        $\mathbf{v} \leftarrow (\boldsymbol{\varepsilon} - \rvz_{full}^{(0)})$\;
        
        $\rvz_{r,e}^{(t)} \leftarrow (1 - t)(\mathbf{z_{r}}^{(0)} \Vert \mathbf{z_{e}}^{(0)}) + t(\boldsymbol{\varepsilon}_{F:2F+L-1})$\;
        $\rvz^{(t)} \leftarrow \mathbf{z_{s}}^{(0)} \Vert \rvz_{r,e}^{(t)}$\;
        
        $\hat{\mathbf{v}} \leftarrow \mathbf{F}_{\bm{\theta}}(\rvz^{(t)}, t, \mathbf{c})$\;
        
        $\mathcal{L} \leftarrow \frac{1}{L+F} \sum_{i=F}^{2F+L-1} \bigl\|\mathbf{v}_i - \hat{\mathbf{v}}_i\bigr\|_2^2$\;
    }
    Update $\bm{\theta}$ using gradients of $\mathcal{L}$\;
}
\end{algorithm}

Given a concatenated full latent sequence
\(\mathbf{z}_{\text{full}}^{(0)}=\operatorname{TemporalConcat}\big(\mathbf{z}_s^{(0)},\mathbf{z}_r^{(0)},\mathbf{z}_e^{(0)}\big)\),
we treat the reasoning+editing block as the generation target during training.  

Given timestep \(t\in[0,1]\) and Gaussian noise \(\boldsymbol{\varepsilon}\sim\mathcal{N}(\mathbf{0},\mathbf{I})\),
we only progressively noise the reasoning and editing parts,
$
\mathbf{z}_{r,e}^{(t)}=(1-t)\big(\mathbf{z}_r^{(0)}\Vert\mathbf{z}_e^{(0)}\big)+t\,\boldsymbol{\varepsilon}_{F:2F+L-1},
$
and form the model input \(\mathbf{z}^{(t)}=\mathbf{z}_s^{(0)}\Vert\mathbf{z}_{r,e}^{(t)}\), 
The target velocity field is \(\mathbf{v}=\boldsymbol{\varepsilon}-\mathbf{z}_{\text{full}}^{(0)}\).
Our model \(\mathbf{F}_{\theta}(\cdot)\) predicts this velocity field from the partially noised input, and we train it by minimizing the mean squared error between predicted and true velocities. Concretely, we only supervise the reasoning and target frames, so the training loss can be written in per-frame form as
\begin{equation}
\label{eq:loss_perframe}
\mathcal{L}
= \frac{1}{L+F}\sum_{i=F}^{2F+L-1}
\Big\|
\mathbf{v}_i
-
\big[\mathbf{F}_{\theta}(\mathbf{z}^{(t)},t,\mathbf{c})\big]_i
\Big\|_2^2,
\end{equation}
where \(\big[\mathbf{F}_{\theta}(\mathbf{z}^{(t)},t,\mathbf{c})\big]_i\) denotes the model's prediction for frame \(i\) and $\mathbf{c}$ is the text condition.
The model parameters \(\mathbf{F}_{\theta}(\cdot)\)  are updated via a gradient step computed from this loss. The full training procedure is summarized in Algorithm~\ref{alg:train}.


During inference we initialize the reasoning+editing block from Gaussian noise,
$\mathbf{z}_{r,e}^{(1)} \sim \mathcal{N}(\mathbf{0},\mathbf{I})$
and form the full latent at \(t=1\) by temporal concatenation with the clean source $
\mathbf{z}_{\text{full}}^{(1)}=\operatorname{TemporalConcat}\big(\mathbf{z}_s^{(0)},\,\mathbf{z}_{r,e}^{(1)}\big).
$
An ODE solver guided by our model \(\mathbf{F}_{\theta}\) evolves \(\mathbf{z}_{\text{full}}^{(t)}\) to \(\mathbf{z}_{\text{full}}^{(0)}\). The source latents \(\mathbf{z}_s^{(0)}\) are held fixed during inference, so only the reasoning/editing parts change.
We then extract the edited-target latent using the same slicing index as in training:
$
\mathbf{z}_{\text{edit}}^{(0)} \;=\; \bigl(\mathbf{z}_{\text{full}}^{(0)}\bigr)_{F+L:2F+L-1}
$
and decode the final edited video:
$
\mathbf{x}_{\text{edit}} \;=\; \vaeDecoder\bigl(\mathbf{z}_{\text{edit}}^{(0)}\bigr).
$



\subsection{Video Data Curation}
\label{sec:data}

The training of our VideoCoF requires a large and diverse dataset structured as source, reasoning, and edited video triplets. 
However, existing video editing datasets and methods predominantly focus on single-instance-level object manipulation.
This limitation is a significant barrier, as real-world videos contain complex visual cues, multiple interacting instances, and intricate spatial relationships (e.g., physical left/right, object-to-object interactions). 
Enabling a generative model to comprehend these complex, instance-level dynamics is a critical step toward true reasoning-based video editing. 
Therefore, we develop a comprehensive data curation pipeline, illustrated in Figure \ref{fig:data_pipe}, to specifically generate and process complex, instance-level video data.

\noindent{\textbf{Instance-Level Curation Pipeline.}}
Our pipeline begins with a large pool of diverse videos sourced from Pexels \citep{pexels}. 
First, we employ the Qwen-VL 72B \citep{Qwen2VL} to perform multi-instance identification, scanning the videos to find scenes that contain multiple, distinct objects. 
Once these videos are identified, we use Grounding-SAM2 \cite{groundedsam} to perform precise segmentation, generating distinct segmentation masks for each individual instance. With these instance-specific masks, we generate triplets for a variety of editing tasks:
\begin{itemize}
    \item \textbf{Object Addition/Removal:} We utilize the Minimaxremover \citep{zi2025minimax} to erase a specific instance from the video. The data for object addition is then created by simply reversing this process.
    
    \item \textbf{Object Swap and Local Style Transfer:} For these tasks, we leverage the VACE-14B \cite{vace} in its inpainting mode to fill the specified masked regions. Critically, the creative prompts for these inpainting edits are generated by GPT-4o\cite{gpt-4o}, as we found Qwen-VL 72B's imaginative capabilities for this specific task to be limited.
\end{itemize}

\noindent\textbf{Filtering and Final Dataset.}
All generated video pairs are rigorously evaluated to ensure quality. We use the Dover Score \cite{dover} to assess aesthetic quality and the VIE Score \citep{viescore} to measure editing fidelity and coherence. 
A weighted combination of these scores is used to filter for high-quality, successful edits. 
Finally, we use this pipeline to filter from the large-scale open-source Señorita 2M \citep{senorita-2m} dataset, and distill a high-quality subset of \textbf{50k} videos to supplement our training data.
This multi-pronged approach yields our final large-scale dataset, rich in the instance-level complexity required for reasoning-based video editing.

\begin{table*}[htbp]
\centering
\caption{
We compare \OursMethod~with SOTA baselines on \OursBench: TokenFlow~\citep{geyer2023tokenflow}, InsV2V~\citep{insv2v}, Señorita~\citep{senorita-2m} (an I2V model guided by InsP2P~\citep{instructpix2pix}), VACE-14B~\citep{vace} (using GPT-4o–generated captions), the concurrent ICVE~\citep{icve} (pretrained on \textbf{1M} videos and fine-tuned on \textbf{150k}), and LucyEdit~\citep{lucyedit}.
Despite the extensive training data used by baselines, \OursMethod~is fine-tuned on only \textbf{50k} video pairs and achieves superior instruction-following and success ratio.}

\resizebox{1.0\textwidth}{!}{
\begin{tabular}{l|cccc|ccc}
\toprule
\textbf{Model} & \multicolumn{4}{c|}{\textbf{GPT-4o Score (avg.)}} & \multicolumn{3}{c}{\textbf{Perceptual Quality (avg.)}} \\
& Instruct Follow$\uparrow$ & Preservation$\uparrow$ & Quality$\uparrow$ & Success Ratio$\uparrow$ & CLIP-T$\uparrow$ & CLIP-F$\uparrow$ & DINO$\uparrow$ \\
\midrule
TokenFlow \citep{geyer2023tokenflow} & 3.12 & 5.85 & 5.10 & 4.25\%  & 25.42 & 0.982 & 0.970 \\
InsV2V \citep{insv2v}                & 3.41 & 6.15 & 5.51 & 6.39\%  & 26.19 & 0.988 & 0.978 \\
Señorita \citep{senorita-2m}         & 3.26 & 6.30 & 5.48 & 10.35\% & 26.04 & 0.994 & 0.988 \\
VACE \citep{vace}                    & 7.47 & 5.82 & 7.61 & 26.60\% & 27.02 & 0.994 & 0.990 \\
ICVE \citep{icve}                    & 7.79 & 8.06 & \textbf{8.14} & 57.76\% & 27.49 & 0.992 & 0.986 \\
Lucy Edit \citep{lucyedit}           & 5.24 & 6.50 & 6.37 & 29.64\% & 26.98 & 0.991 & 0.986 \\

\rowcolor[HTML]{d0ece7}
\textbf{VideoCoF (Ours)}             & \textbf{8.97} & \textbf{8.20} & \underline{7.77} & \textbf{76.36\%} & \textbf{28.00} & 0.992 & 0.991 \\
\bottomrule
\end{tabular}
}
\vspace{-1em}
\label{tab:videocof_bench}
\end{table*}

\section{Experiments}

\begin{figure*}[t!]
\centering
\begin{minipage}[t]{\linewidth}
\centering
\includegraphics[width=1\columnwidth]{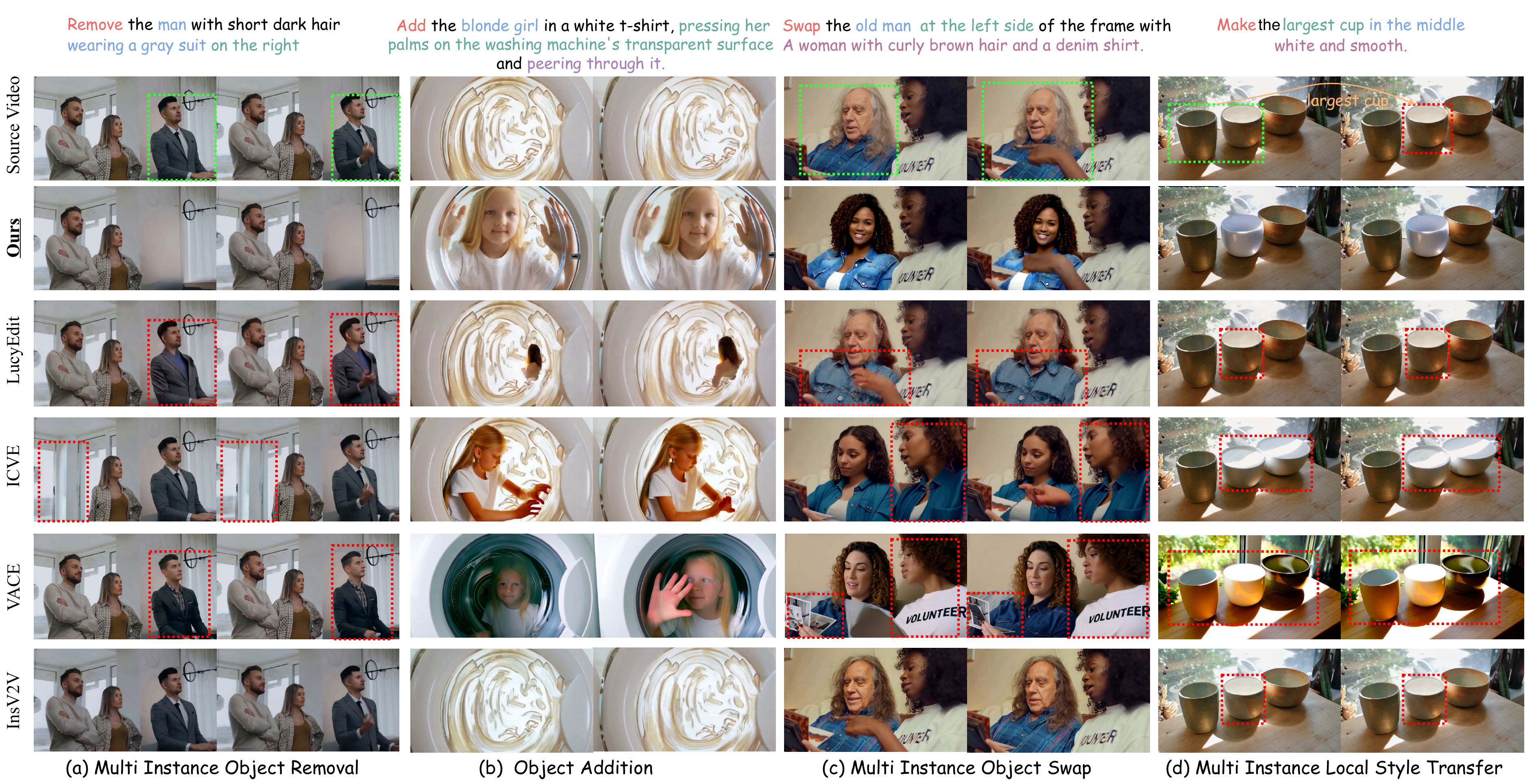}
\end{minipage}
\centering
\caption{Visual comparison between our \OursMethod~and other methods on diverse video editing tasks.}
\label{qualitative_compare} 
\end{figure*}

\begin{table}[h!]
\caption{\textbf{Ablation on Chain of Frames and RoPE design.} 
}
\centering
\small 

\begin{tabular}{lccc}
\toprule

\rowcolor{mygray}
\multicolumn{4}{c}{\textbf{Ablation on Chain of frames and RoPE design}} \\

\rowcolor{mygray}
& \multicolumn{2}{c}{\textbf{Naive Temporal in Context}}
& \multicolumn{1}{c}{\textbf{VideoCoF}} \\

\rowcolor{mygray}
\textbf{CoF}
& \multicolumn{1}{c}{\ding{55}} 
& \multicolumn{1}{c}{\ding{55}} 
& \multicolumn{1}{c}{\textbf{\ding{51}}} \\ 

\rowcolor{mygray}
\textbf{RoPE Design}
& \multicolumn{1}{c}{0--2F-1} 
& \multicolumn{1}{c}{0--F-1, 0--F-1}
& \multicolumn{1}{c}{\textbf{1--F, 0, 1--F}} \\ 

\midrule 

\multicolumn{4}{l}{\textit{GPT-4o Score}} \\
Instruct Follow$\uparrow$ & 8.109 & 8.064 & \textbf{8.973} \\
Preservation$\uparrow$ & 7.930 & 7.793 & \textbf{8.203} \\
Quality$\uparrow$ & 7.394 & 7.217 & \textbf{7.765} \\
Success Ratio$\uparrow$* & 72.41\% & 65.52\% & \textbf{76.36\%} \\

\midrule 

\multicolumn{4}{l}{\textit{Perceptual Quality}} \\
CLIP-T$\uparrow$ & 26.880 & 27.088 & \textbf{28.000} \\
CLIP-F$\uparrow$ & 0.9907 & 0.9905 & \textbf{0.9915} \\
DINO$\uparrow$ & 0.9857 & 0.9826 & \textbf{0.9913} \\
\bottomrule

\end{tabular}

\label{tab:ablation_cof_rope} 
\vspace*{-20pt}
\end{table}

\subsection{Implementation Details.}
\label{sec:implementation_details}

\OursMethod~is trained on WAN-14B \citep{wanx}. We employ a resolution-bucketing strategy to support multiple aspect ratios, using spatial resolutions of 336×592, 400×704, 400×752, and 400×944 (and the corresponding vertical variants, e.g., 592×336). Training videos are sourced from Señorita \citep{senorita-2m} and are 33 frames long. We ultimately train on only \textbf{50k} curated video samples. Thanks to our RoPE alignment design, the model generalizes to longer sequences at inference (e.g., \textbf{513+} frames). 
By default, we use 33-frame source and edited videos, together with a 4-frame reasoning clip.
We train for 8k iterations on 16 H100 GPUs. With DMD-LoRA \citep{yin2024onestep}, inference takes only 4 steps and about \textbf{10 seconds} to edit 33 frames on a single H100 GPU.

\subsection{\OursBench~ and Experimental Setting} 
\noindent\textbf{\OursBench.} Previous video-editing benchmarks such as V2VBench~\citep{v2vbench}, TGVE~\citep{tgve}, and FIVE-Bench~\citep{five} focus on target-prompt edits and are mostly limited to class-level object swaps. They were mainly designed for training-free methods and are not suitable for instruction-guided or instance-level video editing. Real-world editing requires precise instruction understanding, including instance- and part-level control (e.g., distinguishing multiple people or left vs. right), and complex reasoning. To address these gaps, we introduce \OursBench. It contains 200 high-quality videos collected from Pexels \citep{pexels}, covering diverse scenes and both landscape and portrait aspect ratios. \OursBench~includes four tasks: Object Removal, Object Addition, Object Swap, and Local Style Transfer, each with 50 samples. Half of these samples per task are instance-level cases with instance-focused editing prompts.

\noindent\textbf{Evaluation Metrics}. To evaluate editing performance on \OursBench, we employ MLLM-as-a-Judge to provide a holistic evaluation score. This is achieved by prompting \textbf{GPT-4o}~\citep{gpt-4o} to assess multiple criteria given the original video, edited video, and user instruction: (1) Instruction Following (editing accuracy), (2) Preservation (unedited regions), (3) Video Quality. (4) Success ratio: we prompt the GPT-4o to provide a binary Success Ratio (Yes/No) to judge the overall success of the edit. 
We report three perceptual quality metrics to quantify low- and high-level visual similarity between source and target frames: CLIP-T for image–text alignment, CLIP-F for temporal consistency, and DINO for structural consistency.

\subsection{Comparison on \OursBench}

We show qualitative and quantitative comparisons of \OursBench~in this section. 
As shown in Table~\ref{tab:videocof_bench}, we evaluate \OursMethod~against five baseline methods on the \OursBench~ benchmark, which spans four distinct video editing tasks: multi-instance removal, object addition, multi-instance swap, and multi-instance local style transfer. 

Overall, \OursMethod~demonstrates the best performance in \textbf{Instruct Follow} and \textbf{Success Ratio} across all categories. 
Compared to naive temporal in-context editing approaches like ICVE \citep{icve}, our method achieves significantly higher success rates and better instruction adherence using only \textbf{50k
} reasoning pairs, whereas ICVE is pre-trained on 1M samples and fine-tuned on 150k data. 


Qualitatively (see Figure~\ref{qualitative_compare}), our method also shows clearer, more faithful edits at the instance level: (a) Multi-instance removal: we precisely remove the right instance while ICVE\cite{icve} incorrectly removes the left instance.
(b) Object addition: the added girl is correctly placed inside the washing machine, matching the instruction.
(c) Object swap: we replace the elderly person’s face and update clothing; Lucy Edit \cite{lucyedit} changes only clothing, ICVE fails to disambiguate instances, and VACE often alters non-target people.
(d) Local style (multi-instance): our model correctly identifies and edits the largest cup among several similar objects; other methods either fail to edit or mistakenly edit a bowl.
These qualitative examples demonstrate \OursMethod’s stronger instance-level reasoning and higher editing fidelity.

\begin{figure}[t]
    \centering
    \vspace{-10pt}
    \includegraphics[width=\linewidth]{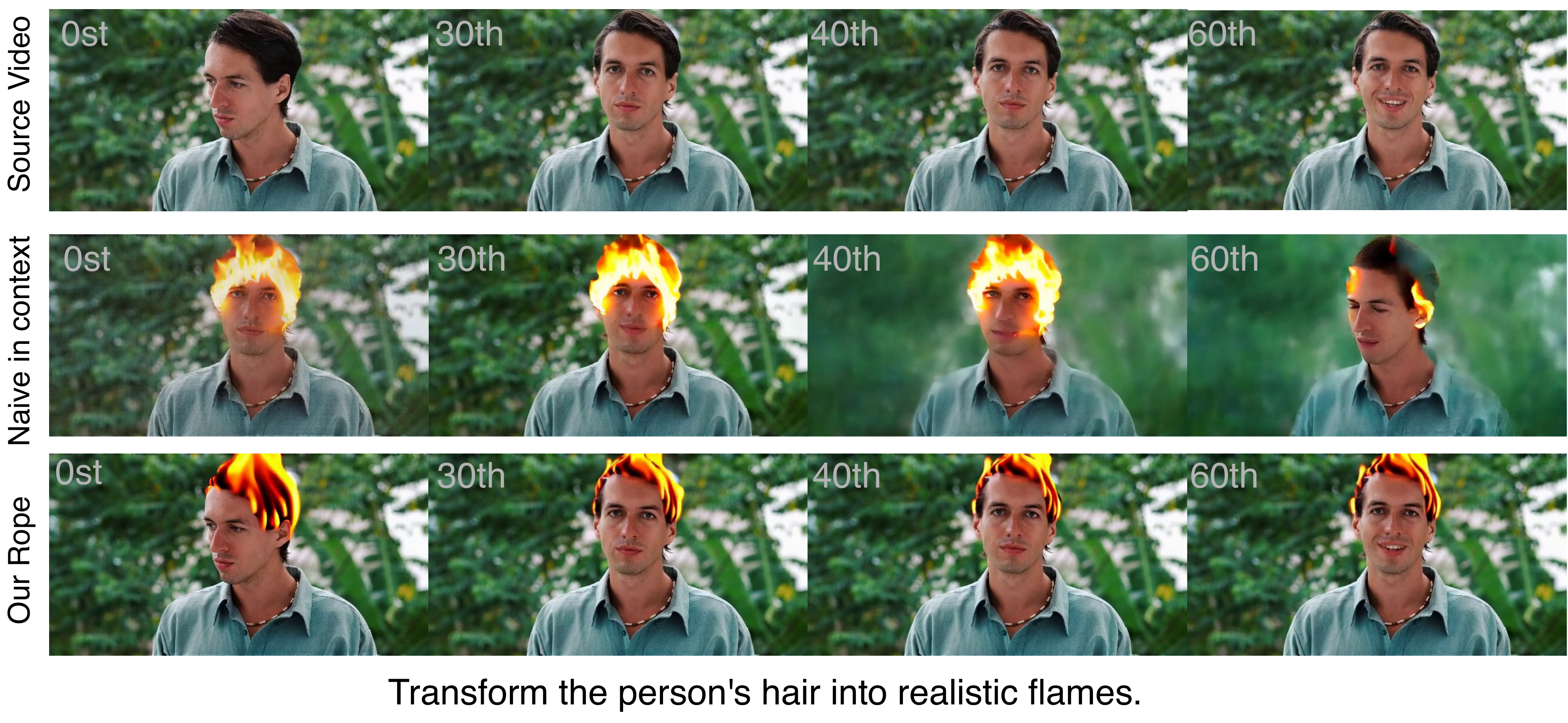}
    \vspace{-15pt}
    \caption{Length extrapolation beyond the training length.}
    \vspace{-5pt}
    \label{fig:length_exploration}
\end{figure}

\subsection{Ablation Study}
\label{sec_abs}

To verify our novel Chain of Frames (CoF) design, particularly its ``reasoning frames" and the RoPE design for length extrapolation, we conduct an ablation study on the reasoning frames, RoPE alignment strategy and reasoning format. 

\noindent \textbf{Naive Temporal Incontext \textit{VS.} CoF.}
As shown in Table~\ref{tab:ablation_cof_rope}, we compare \OursMethod~against a ``Naive Temporal in-context" baseline. 
This applies temporal in-context learning by using the source video as a condition through temporal concatenation, an approach similar to ICVE~\cite{icve}. 

In contrast, our approach introduces \textbf{reasoning frames} as a core component of the (CoF) design. This ensures the video editing follows a reasoning process, i.e., forcing the model to predict the editing region first and then execute the versatile edit within that specific area.

The efficacy of this design is evident when comparing the first ($[0, 2F-1]$) and third (VideoCoF) columns in Table~\ref{tab:ablation_cof_rope}. The inclusion of CoF brings substantial gains: the instruct follow score increases by 10.65\% and the success ratio improves by 5.46\%. Furthermore, the 4.16\% increase in CLIP-T confirms that our reasoning frames effectively enhance the model's editing accuracy and precision.

\noindent \textbf{RoPE Design for Length Extrapolation.}
\label{sec:rope_exploration}
As illustrated in Fig~\ref{fig:length_exploration}, the naive approach ($[0, 2F-1]$) only learns a fixed temporal mapping (e.g., mapping frame $0_{th}$ to frame $33_{th}$). This prevents length extrapolation, causing severe degradation (blurriness, motion misalignment, and artifacts) when a 33-frame trained model is tested on 81 frames (second row).

In contrast, our RoPE alignment design ($[1-F, 0, 1-F]$) generalizes to unseen lengths without quality degradation (third row). As demonstrated in Fig~\ref{fig:teaser}, our model extrapolates to 141 frames (4$\times$ training length) and even 513 frames (16$\times$ training length), supporting strong generalization to much longer video sequences.

This effectiveness is also quantified in Table~\ref{tab:ablation_cof_rope} (third vs. first column). We observe a 3.4\% relative increase in the preservation score. Furthermore, the improved DINO score confirms that our RoPE design better preserves the original video's spatio-temporal structure during editing.

\begin{figure}[t!]
    \centering
    \includegraphics[width=\linewidth]{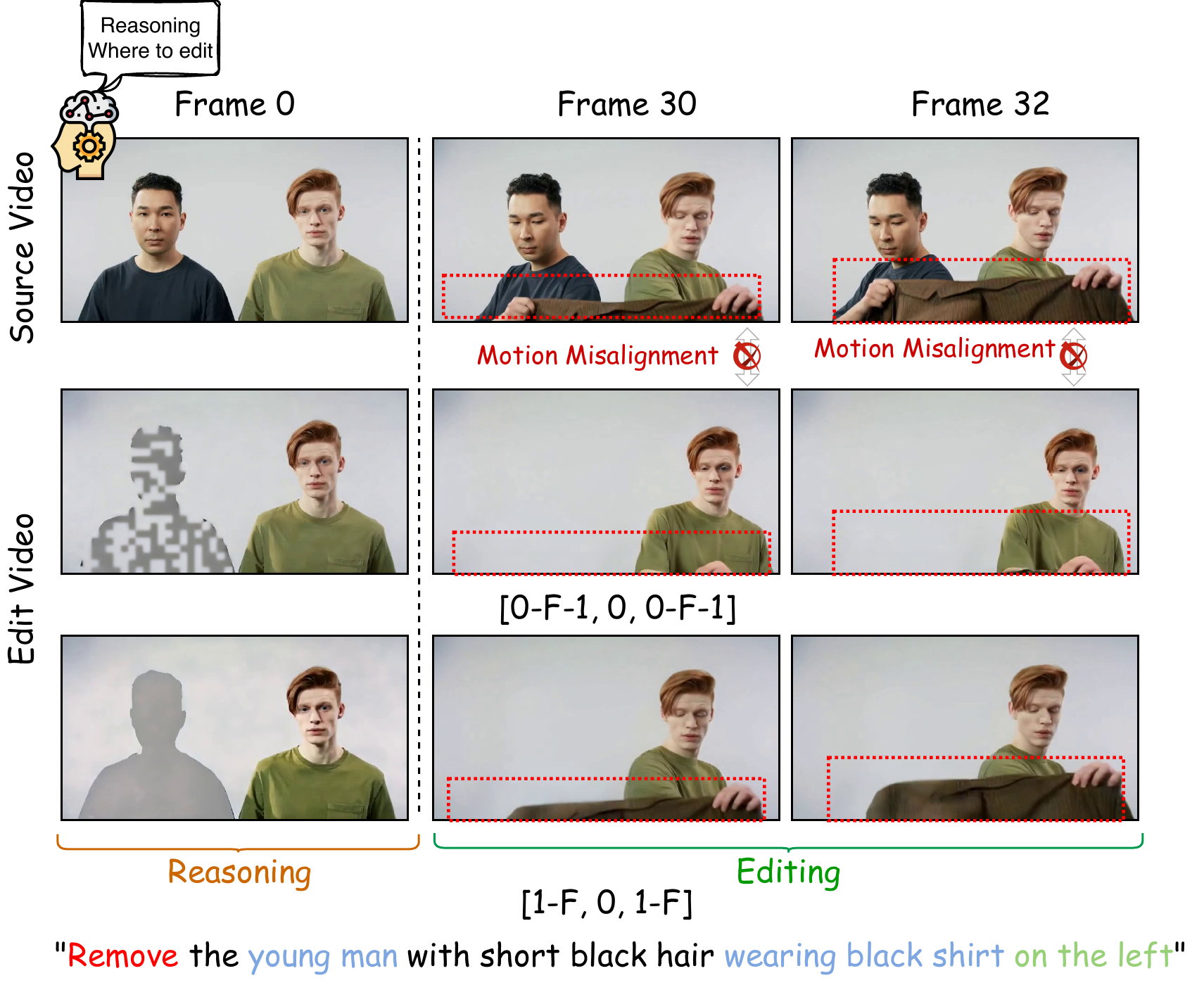}
    \vspace{-20pt}
    \caption{Motion alignment benefit by our RoPE design. \vspace{-10pt}}
    \label{fig:rope_artifact}
\end{figure}

\noindent \textbf{RoPE Design for Motion Alignment.}
Setting the temporal index for the reasoning frame latent is a critical design choice. A naive approach is to set its index to 0, aligning it with the first video frame. This causes two severe issues.

First, it leads to significant motion misalignment (e.g., the subject fails to perform the "lifting clothes" motion in Fig~\ref{fig:rope_artifact}, second row). Second, this "0-index" design causes interference with the first editing video frame (also index 0), leading to artifacts where the model incorrectly predicts the first frame as the reasoning frame (Fig~\ref{fig:reason_vis}).

Therefore, we fix the reasoning latent's index to 0, while the source and edited video indices range from 1 to $F$ (denoted as $[1-F, 0, 1-F]$). This strategy allows the reasoning frame to provide clear spatial guidance on \textbf{where} to edit, without disrupting the video's temporal structure and motion alignment. The improvements across all metrics in Tab~\ref{tab:ablation_cof_rope} (column 3 vs. column 2) validate this design.

\noindent \textbf{Reasoning Frame Format.}
First, we explore the most suitable color for the reasoning frame mask. As shown in Table~\ref{tab:ablation_reason_format}, we compare three formats: (1) A black mask over the unedited region; (2) A red, 50\% transparent highlight, same as Veggie \citep{yu2025veggie}; and (3) A gray, 50\% transparent mask. 
The quantitative results show that using a gray mask (column 3) for the edit region yields the best performance.

Furthermore, we argue that the reasoning frame should act as a gradual transition from the source video to the edited video. Therefore, we test a progressive gray mask. Instead of a single static mask, we interpolate the gray reasoning frame with the editing frame, with the transparency progressively increased (e.g., 0\%, 25\%, 50\%, 75\%). As shown by comparing column 4 and column 3 in Table~\ref{tab:ablation_reason_format}, this progressive gray reasoning frame approach works best.

\begin{figure}[htbp]
    \vspace{-10pt}
    \centering
    \includegraphics[width=\linewidth]{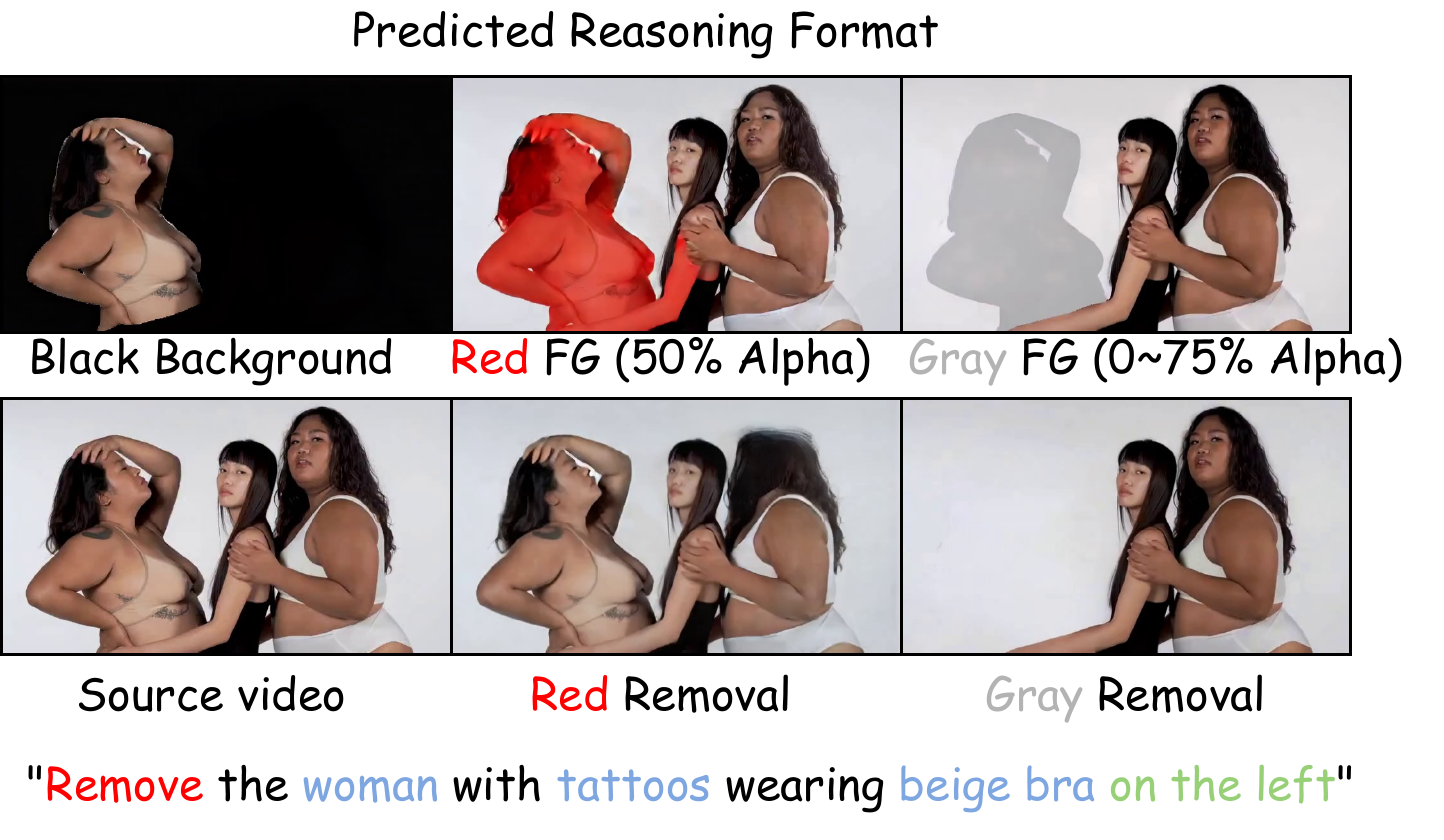}
    \vspace{-2em}
    \caption{Ablation on reasoning frame format. \vspace{-15pt}}
    \label{fig:reason_format}
\end{figure}

Qualitatively, as shown in Figure~\ref{fig:reason_format}, the mask format is critical. The black mask fails the deletion task, while the red mask incorrectly deletes content on the right side. In contrast, our progressive gray mask accurately performs the intended deletion on the left. We conclude from these experiments that the optimal reasoning format is a gray mask with progressive transparency. 

\begin{table}[t!]
\caption{\textbf{Ablation on the reasoning frame format.}}
\centering
\small 

\sisetup{table-align-text-post=false}
\begin{tabular}{
    l 
    S[table-format=2.3] 
    S[table-format=2.3] 
    S[table-format=2.3] 
    S[table-format=2.3] 
}
\toprule

\rowcolor{mygray}
\multicolumn{5}{c}{\textbf{Ablation on Reasoning Frame Format}} \\

\rowcolor{mygray}
\textbf{Color} 
& {\textbf{Black (bg)}} 
& {\textbf{Red}}     
& {\textbf{Gray}}    
& {\textbf{Gray}} \\

\rowcolor{mygray}
\textbf{Transparency} 
& {(0\%)}              
& {(50\%)}              
& {(50\%)}           
& {\textbf{(0-75\%)}} \\ 

\midrule 

\multicolumn{5}{l}{\textit{GPT-4o Score}} \\

Instruct Follow$\uparrow$ & 7.512 & 7.805 & 8.150 & \textbf{8.973} \\ 
Preservation$\uparrow$    & 7.034 & 7.350 & 7.443 & \textbf{8.203} \\ 
Quality$\uparrow$         & 6.155 & 6.501 & 6.645 & \textbf{7.765} \\
Success Ratio$\uparrow$* & 52.17\% & 60.33\% & 68.45\% & \textbf{76.36\%} \\ 

\midrule

\multicolumn{5}{l}{\textit{Perceptual Quality}} \\ 
CLIP-T$\uparrow$ & 26.550 & 26.810 & 27.220 & \textbf{28.000} \\
CLIP-F$\uparrow$ & 0.9810 & 0.9855 & 0.9889 & \textbf{0.9915} \\
DINO$\uparrow$   & 0.9750 & 0.9790 & 0.9803 & \textbf{0.9913} \\
\bottomrule
\end{tabular}
\vspace{-10pt}
\label{tab:ablation_reason_format}
\end{table}

\section{Conclusion}
In this paper, we introduced \OursMethod, a unified model for universal video editing via temporal reasoning. 
We identified that existing temporal in-context learning approaches often fail due to a lack of explicit spatial cues, leading to weak instruction-to-region mapping and imprecise localization.
To address these issues, we proposed the innovative Chain of Frames.
VideoCoF compels the video diffusion model to follow a ``see, reason, then edit" process by first predicting the editing region before executing the versatile edit. 
Furthermore, to solve the length generalization challenge, we developed a novel RoPE alignment paradigm that accounts for the reasoning latent.
This design enables up to 16$\times$ length extrapolation during the inference. Experimental results show that \OursMethod~achieves SOTA performance using only 50k video pairs, validating the efficiency and effectiveness of our temporal reasoning design.

{
    \small
    \bibliographystyle{ieeenat_fullname}
    \bibliography{main}
}


\clearpage
\setcounter{page}{1}
\maketitlesupplementary

This document provides more details of our approach and additional experimental results, which are organized as follows:
\begin{itemize}
\setlength{\itemsep}{0pt}
    \item Discussion on RoPE Design (\S\ref{sec:rope_unic_difference})
    \item Editing Length Upper Bound (\S\ref{sec:editing_upper_bound})
    \item Full Comparison (\S \ref{sec:full_comparison})
    \item TGVE+ and V2VBench (\S\ref{sec:tgve_v2vbench})
    \item More Ablation studies (\S\ref{sec:more_ablation})
    \item Implementation Details (\S\ref{sec:implementation})
    \item Metrics (\S\ref{sec:metrics})
    \item Future Directions (\S\ref{sec:future_directions})
\end{itemize}

\section{Discussion on RoPE Design}
\label{sec:rope_unic_difference}

\noindent\textbf{Difference from UNIC.}
VideoCoF targets \emph{length extrapolation} in V2V editing, whereas UNIC focuses on \emph{source-target alignment}. To enable extrapolation, we \textbf{pre-position} the ID/reasoning frames so that their indices never overlap with the target video. In contrast, UNIC \textbf{post-positions} the ID/reasoning tokens (e.g., shifting them to a later index range). As the video becomes longer and reaches the same index range, overlap becomes unavoidable, leading to temporal index collisions and content perturbations.

\begin{figure}[htbp]
    \centering
    \vspace{-1em}
    \includegraphics[width=1.0\linewidth]{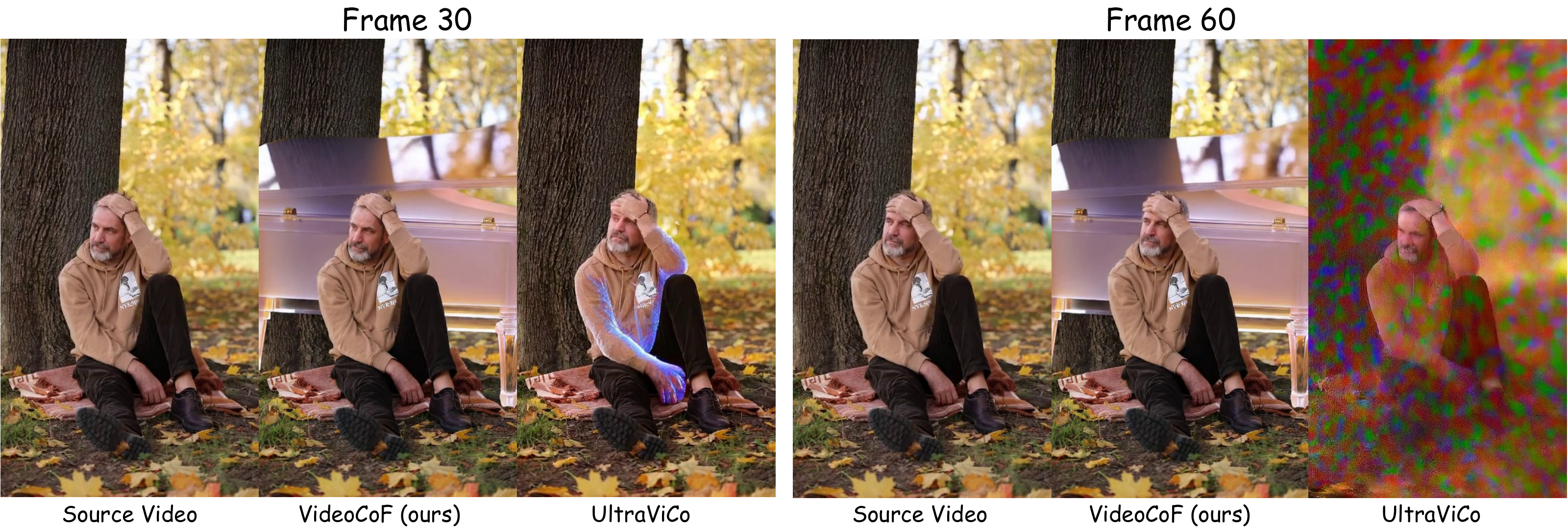}
    \vspace{-1em}
    \caption{Comparison between VideoCoF and UltraViCo}
    \label{fig:videocof_vs_ultravico}
    \vspace{-1em}
\end{figure}

\noindent\textbf{Difference from T2V extrapolation methods.}
RIFLEx~\citep{zhao2025riflex} and UltraViCo~\citep{zhao2025ultravico} mainly address motion or content repetition in \emph{T2V} length extrapolation, whereas our \emph{V2V} extrapolation is primarily limited by source-target alignment and temporal index collision. We apply UltraViCo to the baseline, and Fig.~\ref{fig:videocof_vs_ultravico} shows that it still fails.

\section{Editing Length Upper Bound}
\label{sec:editing_upper_bound}

\begin{figure}[htbp]
    \centering
    \vspace{-1em}
    \includegraphics[width=1.0\linewidth]{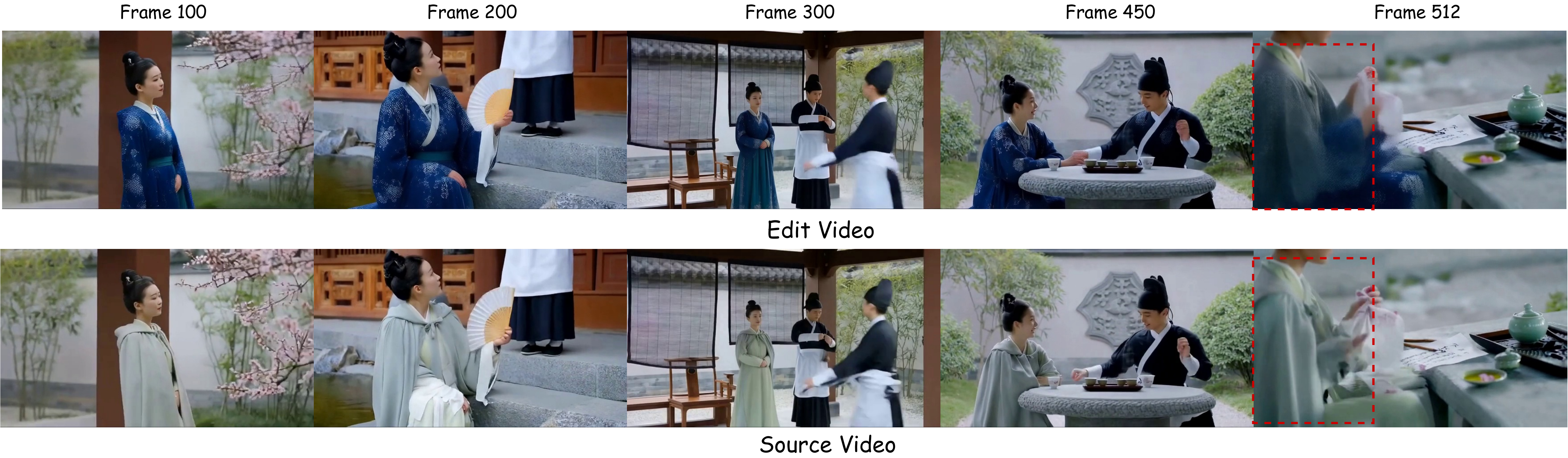}
    \caption{Temporal extrapolation failure case.}
    \label{fig:temporal_extrapolation_512}
\end{figure}

\noindent \textbf{Systematic Temporal Extrapolation Failure Cases.}
We test $16\times$ extrapolation in both single-shot and multi-shot settings. Single-shot setting supports $16\times$ extrapolation. For multi-shot, as shown in Fig.~\ref{fig:temporal_extrapolation_512}, it remains stable for the first 500 frames but shows slight degradation at 512 frames.

\section{Full Comparison}
\label{sec:full_comparison}
As shown in Tab.~\ref{tab:full_comparison}, we provide a detailed breakdown of the results across four distinct tasks: Object Removal, Object Addition, Object Swap, and Local Style Transfer. Our \OursMethod~consistently achieves the highest scores in \textbf{instruction following} and \textbf{success ratio} across all tasks, demonstrating superior capability in understanding and executing editing requests. We note that our scores in Preservation and Quality are slightly lower than the concurrent work ICVE \citep{icve}. This performance gap is reasonable given that ICVE benefits from large-scale pre-training on 1M pairs, and its supervised fine-tuning (SFT) dataset scale (150K) is three times larger than ours (50k).
Furthermore, in terms of perceptual quality, \OursMethod~achieves the highest \textbf{CLIP-T} scores across all tasks. This further demonstrates superior video-text alignment, consistent with our leading performance in GPT-4o score.

\section{TGVE+ and V2VBench}
\label{sec:tgve_v2vbench}
Since TGVE\citep{wu2023cvpr} has only 304 samples, we instead evaluate on the comprehensive TGVE+ subset used in EVE \citep{eve} (1,417 samples across 7 instruction editing tasks), and report V2VBench \citep{v2vbench} results. As shown in the tables, \OursMethod~achieves the best performance on both benchmarks.

\begin{table}[h]
\centering
\caption{Comparison on the TGVE+ benchmark.}
\resizebox{1.0\columnwidth}{!}{
\setlength\tabcolsep{4pt}
\renewcommand\arraystretch{1.1}
\begin{tabular}{l|l||cccc}
\toprule

\rowcolor[HTML]{F2F2F2}
\textbf{Dataset} & \textbf{Methods} & \textbf{PickScore $\uparrow$} & \textbf{CLIP-F $\uparrow$} & \textbf{ViCLIP$_{dir} \uparrow$} & \textbf{ViCLIP$_{out} \uparrow$} \\

\hline\hline 

\multirow{6}{*}{\textbf{TGVE+}} 
& Tune-A-Video ~\citep{tuneavideo}                  & 20.47 & 0.933 & 0.131 & 0.242 \\
& SDEdit ~\citep{sdedit}                        & 20.35 & 0.899 & 0.131 & 0.241 \\
& STDF  ~\citep{stdf}                         & 20.60 & 0.933 & 0.093 & 0.227 \\
& Fairy ~\citep{fairy}                          & 19.81 & \underline{0.933} & 0.140 & 0.197 \\
& InsV2V ~\citep{insv2v}                         & 20.37 & 0.925 & 0.174 & 0.236 \\ 
& EVE  ~\citep{eve}                          & \underline{20.88} & 0.926 & \underline{0.198} & \underline{0.251} \\

\midrule

\rowcolor{gray!20}
& \textbf{VideoCoF}       & \textbf{20.90} & \textbf{0.956} & \textbf{0.213} & \textbf{0.257} \\ 

\bottomrule
\end{tabular}
}
\vspace{-4mm}
\end{table}
\begin{table}[htbp]
\centering
\caption{Comparison on the V2VBench benchmark.}
\resizebox{1.0\columnwidth}{!}{
\setlength\tabcolsep{2pt}
\renewcommand\arraystretch{1.1}
\begin{tabular}{l||ccccccc}
\toprule
\rowcolor[HTML]{F2F2F2}
\textbf{V2VBench} & \begin{tabular}[c]{@{}c@{}}\textbf{Frames}\\ \textbf{Quality} $\uparrow$\end{tabular} & \begin{tabular}[c]{@{}c@{}}\textbf{Semantic}\\ \textbf{Consis.} $\uparrow$\end{tabular} & \begin{tabular}[c]{@{}c@{}}\textbf{Object}\\ \textbf{Consis.} $\uparrow$\end{tabular} & \begin{tabular}[c]{@{}c@{}}\textbf{Frames-Text}\\ \textbf{Align.} $\uparrow$\end{tabular} & \begin{tabular}[c]{@{}c@{}}\textbf{Frames}\\ \textbf{Pick.} $\uparrow$\end{tabular} & \begin{tabular}[c]{@{}c@{}}\textbf{Video-Text}\\ \textbf{Align.} $\uparrow$\end{tabular} & \begin{tabular}[c]{@{}c@{}}\textbf{Motion}\\ \textbf{Align.} $\uparrow$\end{tabular} \\
\hline\hline

Tune-A-Video~\citep{tuneavideo}   & \underline{5.001} & 0.934 & 0.917 & 27.513 & 20.701 & 0.254 & -5.599 \\
SimDA~\citep{xing2024simda}          & 4.988 & 0.940 & 0.929 & 26.773 & 20.512 & 0.248 & -4.756 \\
VidToMe~\citep{li2024vidtome}       & 4.988 & 0.949 & 0.945 & 26.813 & 20.546 & 0.240 & -3.203 \\
VideoComposer~\citep{wang2023videocomposer}  & 4.429 & 0.914 & 0.905 & \underline{28.001} & 20.272 & \underline{0.262} & -8.095 \\
MotionDirector~\citep{zhao2024motiondirector}  & 4.984 & \underline{0.949} & \underline{0.951} & 27.845 & \underline{20.923} & \underline{0.262} & \underline{-3.088} \\
\midrule
\rowcolor{gray!20}
\textbf{VideoCoF (Ours)} & \textbf{5.024} & \textbf{0.954} & \textbf{0.956} & \textbf{28.336} & \textbf{21.154} & \textbf{0.275} & \textbf{-2.620} \\
\bottomrule
\end{tabular}
}
\end{table}

\section{More Ablation Studies}
\label{sec:more_ablation}
In this section, we validate key design choices of \OursMethod: the length of reasoning frames and the dispatch prompt.

\subsection{Ablation on Reasoning Frames}

\begin{table}[htbp]
\caption{\textbf{Ablation on the number of Reasoning Frames.} We investigate the impact of varying the number of reasoning frames from 1 to 5. Our default setting (4 frames) achieves the best balance.}
\centering
\small 

\setlength{\tabcolsep}{1.8pt}

\begin{tabular}{l||ccccc}
\toprule

\rowcolor{mygray}
& \multicolumn{5}{c}{\textbf{Ablation on Reasoning Frames}} \\

\rowcolor{mygray}
\textbf{Frames} & 1 & 2 & 3 & \textbf{4 (Ours)} & 5 \\

\hline\hline

\textit{GPT-4o Score} & & & & & \\ 

Instruct Follow$\uparrow$ & 8.219 & 8.312 & 8.281 & \textbf{8.973} & 7.915 \\

Preservation$\uparrow$    & 8.115 & 8.150 & 8.191 & \textbf{8.203} & 6.542 \\

Quality$\uparrow$         & 7.692 & 7.752 & 7.735 & \textbf{7.765} & 5.274 \\

Success Ratio$\uparrow$   & 68.47\% & 69.39\% & 68.32\% & \textbf{76.36\%} & 29.06\% \\

\midrule 

\textit{Perceptual Quality} & & & & & \\
CLIP-T$\uparrow$ & 27.092 & 27.148 & 27.136 & \textbf{28.000} & 26.997 \\
CLIP-F$\uparrow$ & 0.9892 & 0.9893 & 0.9899 & \textbf{0.9915} & 0.9849 \\
DINO$\uparrow$   & 0.9815 & 0.9827 & 0.9836 & \textbf{0.9913} & 0.9719 \\

\bottomrule
\end{tabular}

\label{tab:ablation_reasoning_frames}
\vspace*{-10pt}
\end{table}

Tab.~\ref{tab:ablation_reasoning_frames} investigates the optimal number of reasoning frames ($F$) for spatial guidance. Considering the VideoVAE temporal compression formula $L = (F - 1) // 4 + 1$, frames $1\sim4$ map to a single latent frame ($L=1$), while $F=5$ introduces latent frames $L=2$.
Results show that $F=4$ achieves the best performance. This indicates maximizing spatial information within a single latent frame is more effective than expanding to a second latent, which introduces unnecessary temporal complexity and degradation.

\subsection{Ablation on Temporal Triptych Prompt}
\begin{figure}[t!]
    \centering
    \includegraphics[width=\linewidth]{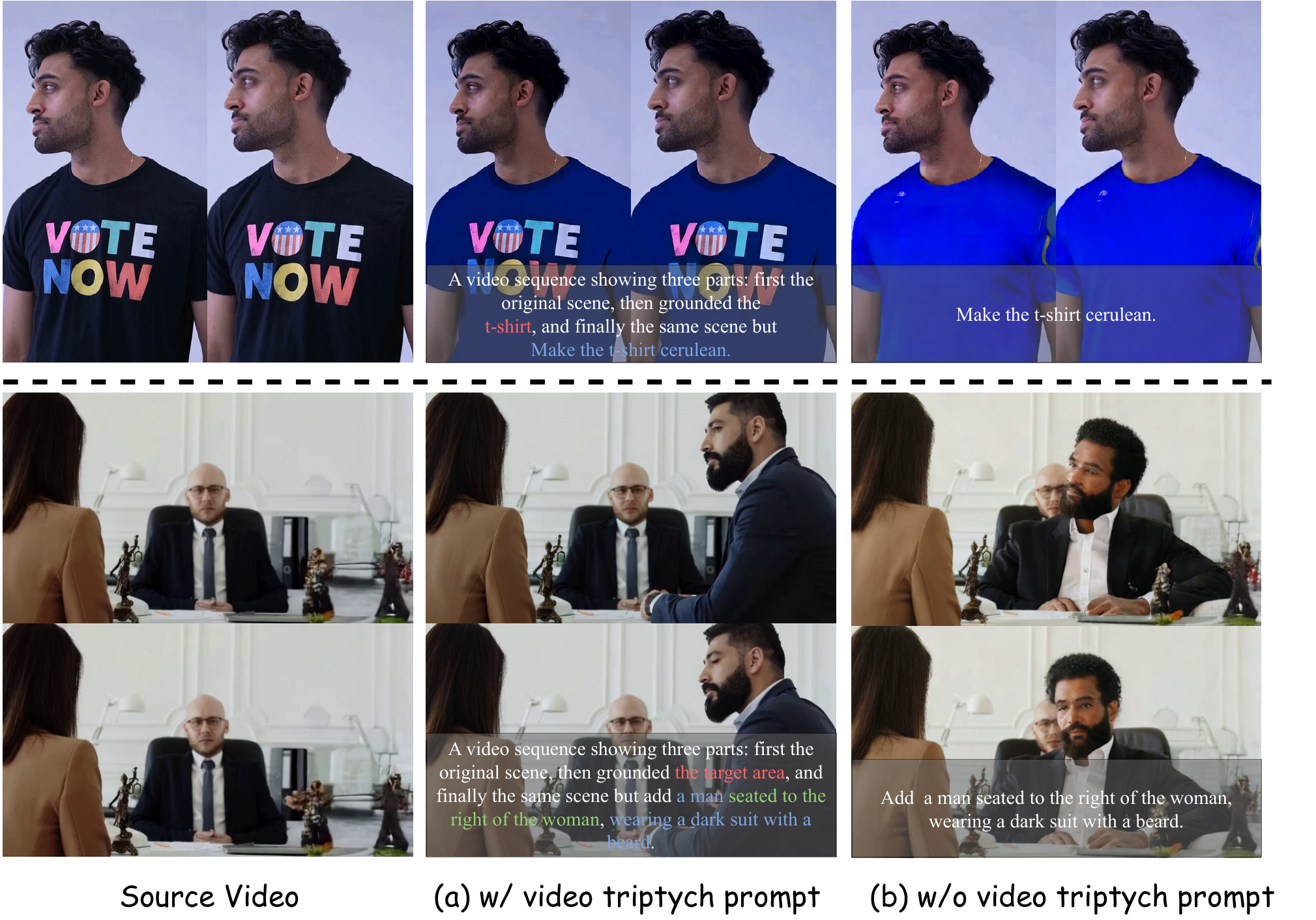}
    \vspace{-10pt}
    \caption{Input Prompt Variants for In-Context Video Editing. We evaluate two prompt formats: (a) Temporal Triptych Prompt - instructions embedded in a structure "A video sequence showing three parts: first the original scene, then grounded \{ground instruction\}, and finally the same scene but \{edit instruction\}.")  (b) Direct Instruction - explicit editing commands provided directly. \vspace{-10pt}}
    \label{fig:triptych_prompt}
\end{figure}
\begin{table}[htbp]
\caption{\textbf{Ablation on Temporal Triptych Prompt.} We compare the performance of our model with and without the triptych patch prompt mechanism. The inclusion of the triptych prompt significantly enhances instruction following and overall success rates.}
\centering
\small

\begin{tabular}{l||cc}
\toprule

\rowcolor{mygray}
& \multicolumn{2}{c}{\textbf{Ablation on Temporal Triptych Prompt}} \\

\rowcolor{mygray}
 & \textbf{w/o Triptych } & \textbf{w/ Triptych (Ours)} \\

\hline\hline

\textit{GPT-4o Score} & & \\
Instruct Follow$\uparrow$ & 8.064 & \textbf{8.973} \\
Preservation$\uparrow$ & 8.094 & \textbf{8.203} \\
Quality$\uparrow$ & 7.360 & \textbf{7.765} \\
Success Ratio$\uparrow$ & 71.43\% & \textbf{76.36\%} \\

\midrule

\textit{Perceptual Quality} & & \\
CLIP-T$\uparrow$ & 27.07 & \textbf{28.00} \\
CLIP-F$\uparrow$ & 0.989 & \textbf{0.992} \\
DINO$\uparrow$ & 0.980 & \textbf{0.991} \\

\bottomrule
\end{tabular}

\label{tab:ablation_dispatch_prompt}
\vspace*{-10pt}
\end{table}

To adapt a standard T2V model for instruction-based editing tasks, we draw inspiration from in-context image editing approaches \citep{iclora,diptychprompting,icedit}. Specifically, we implement a \textbf{temporal triptych prompt} mechanism in \OursMethod~to describe the evolution of video content along the temporal dimension. 
As illustrated in Fig.~\ref{fig:triptych_prompt}, our prompt template is structured as follows: \textit{``A video sequence showing three parts: first the original scene, then grounded \{ground instruction\}, and finally the same scene but \{edit instruction\}.''}

As evidenced in Tab.~\ref{tab:ablation_dispatch_prompt}, this mechanism brings significant performance gains across all metrics. Crucially, unlike the concurrent work ICVE \citep{icve}, which requires computationally expensive pre-training on 1M video pairs to align the T2V model with an instruction mode, our ``temporal triptych prompt" approach offers a practically zero-cost solution to effectively bridge the gap between generation and editing without the need for massive instruction tuning.


\section{Implementation Details}
\label{sec:implementation}
\subsection{Training Dataset}
\begin{table}[t]
\centering
\small 
\setlength{\tabcolsep}{4pt} 
\caption{\textbf{Statistics of the \OursMethod~Training Data.} The dataset consists of 50k samples balanced across four tasks.}
\begin{tabular}{l c p{4.5cm}} 
\toprule
\textbf{Dataset} & \textbf{\#Samples} & \textbf{Information} \\ 
\midrule

\rowcolor{gray!10}\multicolumn{3}{c}{\textbf{Video Editing Tasks}} \\ 
\midrule

\textbf{Obj. Addition} & 10,000 & 
Derived from filtered Señorita. Source generated by removing objects from target via MiniMax-Remover \citep{zi2025minimax}. (absent $\to$ present). \\ \midrule

\textbf{Obj. Removal} & 15,000 & 
Derived from filtered Señorita. Target generated via MiniMax-Remover \citep{zi2025minimax}. Includes 5k multi-instance samples. (present $\to$ absent). \\ \midrule

\textbf{Obj. Swap} & 15000 & 
Generated via VACE-14B \cite{vace} using GPT-4o prompts and Grounding DINO masks. Covers rigid \& non-rigid swaps and 5k multi-instance object swap samples. \\ \midrule

\textbf{Local Style} & 10000 & 
Generated via VACE-14B \cite{vace} using GPT-4o prompts and Grounding DINO masks. Focuses on texture \& stylization. \\ \midrule

\textbf{Total} & \textbf{50,000} & \textbf{Unified Dataset} \\ 

\bottomrule
\end{tabular}

\vspace{-2em}
\label{tab:video_editing_data_statistics}
\end{table}
To equip our model with robust instruction-following capabilities, we constructed a unified chain-of-frames video editing dataset comprising 50k video pairs. As detailed in Table~\ref{tab:video_editing_data_statistics}, the dataset is strategically balanced across four core editing tasks: object addition, removal, swapping, and local stylization. The data construction pipeline integrates both filtered open-source data and high-quality synthetic data. Object Addition and Removal: These subsets (25k samples total) are derived from the Señorita dataset. We employ MiniMax-Remover~\citep{zi2025minimax} to synthesize paired data. Specifically, for the removal task (15k), we treat the original video as the source and the object-erased version as the target. Conversely, for the addition task (10k), we invert this pair (absent $\to$ present). Notably, the removal subset includes 5,000 samples featuring multi-instance objects to enhance model robustness in complex scenes.

Object Swap and Local Style: To capture fine-grained structural and textural changes, we generated 25,000 samples (15k for swap, 10k for style) utilizing VACE-14B~\citep{vace}. The generation process is guided by GPT-4o for diverse prompt synthesis and Grounding DINO for precise mask extraction. The swap subset encompasses both rigid and non-rigid object replacements, while the local style subset focuses on texture modification and artistic stylization.

\subsection{VideoCoF-Bench}

\noindent \textbf{Benchmark Construction.} 
To strictly evaluate the generalization capability, we introduce \OursBench, a diverse evaluation set specifically curated to have no overlap with the training domain. The benchmark is constructed from three distinct sources to ensure comprehensive coverage:

\begin{itemize}
    \item \textbf{Pexels \citep{pexels} Subset:} We manually curated a collection of high-quality videos from Pexels, comprising 50 samples for each editing task. These samples are balanced across the four core editing tasks (Addition, Removal, Swap, and Local Style) to test resolution adaptability and instruction following in varied scenes.
    
    \item \textbf{Standard Benchmark Integration:} To ensure a fair comparison with existing methods, we incorporated representative samples from established benchmarks, including EditVerse~\cite{editverse} and UNIC-Bench~\cite{unic}.
    
    \item \textbf{Adaptation for Fairness:} Notably, for samples sourced from UNIC-Bench (which typically involves ID-driven editing), we removed the reference identity images. This adaptation unifies the evaluation protocol, focusing purely on text-driven editing capabilities. 
\end{itemize}

This combination results in a highly diverse benchmark that challenges models with unseen content and complex editing instructions.

\section{Metrics}
\label{sec:metrics}
\noindent \textbf{GPT Evaluation.} 
To comprehensively assess the editing performance, we employ the state-of-the-art Vision-Language Model, GPT-4o~\citep{gpt-4o}, serving as an automated judge. Following the protocol of InstructX~\citep{instructx}, we sample three frames from each video pair and utilize structured prompts in ~\citep{instructx} to evaluate the results across the following dimensions:

\begin{itemize}
    \item \textbf{Instruction Following (Score 1-10):} This metric measures the precision with which the edit adheres to the user's specific command. Higher scores indicate that the editing result strictly follows the prompt instructions without ambiguity.

    \item \textbf{Visual Quality (Score 1-10):} This evaluates whether the edited video is visually seamless, natural-looking, and aesthetically pleasing. It penalizes artifacts, distortions, or unnatural transitions introduced during the editing process.

    \item \textbf{Preservation (Score 1-10):} This assesses the coherence with the original video context. It strictly penalizes unintended changes to non-edited regions, ensuring the background and non-target objects remain intact.

    \item \textbf{Success Rate (Binary Yes/No):} To mitigate scoring variance, we incorporate a stricter discrete metric inspired by~\citep{icve}. GPT-4o performs a binary judgment based on a rigorous three-step verification logic: (1) \textit{Target Identification} (confirming the target matches the descriptor/position); (2) \textit{Modification Accuracy} (verifying the specific edit is applied); and (3) \textit{Strict Preservation} (ensuring no other instances are altered).
\end{itemize}

As presented in Table~\ref{tab:full_comparison}, \OursMethod~achieves superior performance across all these metrics, validating the effectiveness of our reasoning-driven approach.

\vspace{4pt}
\noindent \textbf{Perception Quality.} 
In addition to semantic evaluation, we report quantitative metrics to measure the visual alignment and temporal consistency:

\begin{itemize}
    \item \textbf{CLIP-T (Text-Image Alignment):} This metric assesses the semantic alignment between the editing instruction and the output video. We compute the cosine similarity between the CLIP~\citep{clip} text embedding of the instruction and the CLIP vision embedding of each output frame, reporting the average score across all frames.
    
    \item \textbf{CLIP-F (Frame-wise Consistency):} To evaluate temporal stability, we utilize the ViT-L/14 vision encoder from CLIP to extract features for each frame. The consistency score is calculated as the average cosine similarity between feature vectors of adjacent frames.
    
    \item \textbf{DINO (Structure Consistency):} While CLIP focuses on semantics, we aim to capture more fine-grained structural and textural consistency. We repeat the temporal consistency calculation using features extracted from a pre-trained DINOv2~\citep{dino} model. DINO's self-supervised training enables it to capture object-level details that might be overlooked by CLIP.
\end{itemize}

\section{Future Directions}
\label{sec:future_directions}

\noindent \textbf{Scaling up Chain-of-Frames.}
Currently, \OursMethod~achieves SOTA performance in instruction following and success rate using only 50k source-reasoning-editing pairs. This demonstrates remarkable data efficiency compared to existing large-scale baselines. For instance, EditVerse~\citep{editverse} utilizes 4M videos and 8M images, ICVE~\citep{icve} leverages 2M pre-training data with 150k SFT samples, and InstructX~\citep{instructx} employs 200k SFT samples with joint training.
Despite the significant gap in data scale, our method's superior performance suggests that the ``reasoning-then-editing'' paradigm is highly effective for Video Diffusion Models (VDMs).
A promising future direction is to explore the performance ceiling of VideoCoF by scaling the dataset to 200k or even millions of samples. Investigating how the reasoning capabilities evolve with larger-scale data could reveal new upper limits for precise video editing.

\noindent \textbf{Joint Image-Video Editing and Efficient Architectures.}
While our current work focuses on video data, integrating high-quality image editing datasets (e.g., MagicBrush \cite{magicbrush}, NHR-Edit\cite{nhredit}) presents a valuable opportunity. Many recent studies have shown that joint training can enhance visual quality and concept understanding.
Future work could investigate the optimal mixture ratios between image and video datasets to maximize performance. Furthermore, designing unified and efficient attention mechanisms is crucial for handling the varying temporal dimensions of images and videos within a single model. Such advancements would likely improve the model's cross-modal learning capabilities, allowing it to transfer fine-grained editing skills from images to complex video dynamics.

\noindent \textbf{Generalizing VideoCoF to Broader Tasks.}
\OursMethod~has demonstrated exceptional performance in local editing tasks. However, the underlying reasoning framework is inherently flexible and can be extended to a wider range of applications.
For \textit{Global Editing} (e.g., style transfer), the reasoning frame could employ a full-frame gray mask to guide global transformations.
For \textit{ID-Driven Editing}, reference identity images could be integrated as ``reasoning frames'' to guide specific character insertions or swaps.
Unifying these diverse tasks—ranging from local modifications to global stylization and ID injection—under the VideoCoF paradigm represents an exciting avenue for future exploration.

\begin{table*}[t!]
\centering
\caption{\textbf{Quantitative full comparison over 4 video editing tasks on \OursBench}. We compare \OursMethod~with SOTA baselines: InsV2V \citep{insv2v}; Señorita \citep{senorita-2m} (an I2V model guided by an InstructPix2Pix \citep{instructpix2pix} first frame); VACE-14B \citep{vace} (using GPT-4o generated captions); the concurrent work ICVE \citep{icve} (pre-trained 1M, fine-tuned 150k); and Lucy Edit Dev \citep{lucyedit}. Despite extensive baseline training data, our \OursMethod~ is fine-tuned on only 50k source-reasoning-editing triplets and shows superior instruction following and success ratio.}

\resizebox{\textwidth}{!}{
\begin{tabular}{l|cccc|ccc}
\toprule
\textbf{Model} & \multicolumn{4}{c|}{\textbf{GPT-4o Score}} & \multicolumn{3}{c}{\textbf{Perceptual Quality}} \\
& Instruct Follow$\uparrow$ & Preservation$\uparrow$ & Quality$\uparrow$ & Success Ratio$\uparrow$ & CLIP-T$\uparrow$ & CLIP-F$\uparrow$ & DINO$\uparrow$ \\
\midrule

\multicolumn{8}{c}{\textbf{Object Removal}} \\
\midrule
InsV2V \citep{insv2v}      & 3.11 & 4.02 & 3.77 & 3.92\%  & 26.85 & 0.984 & 0.973 \\
Señorita \citep{senorita-2m}    & 3.11 & 4.68 & 4.38 & 9.80\%  & 26.96 & \underline{0.995} & 0.990 \\
VACE \citep{vace}        & N/A  & N/A  & N/A  & 0.00\%  & 25.57 & \textbf{0.996} & \underline{0.995} \\
ICVE \citep{icve}        & \underline{5.38} & \underline{7.30} & \textbf{7.68} & \underline{25.49\%} & 26.64 & 0.994 & 0.989 \\
Lucy Edit \citep{lucyedit}   & 2.06 & 4.09 & 4.45 & 1.96\%  & \underline{27.37} & 0.992 & 0.988 \\

\rowcolor[HTML]{e8daef}
\textbf{VideoCoF (Ours)} & \textbf{9.65} & \textbf{7.35} & \underline{6.94} & \textbf{86.27\%} & \textbf{27.50} & 0.988 & \textbf{0.996} \\
\midrule

\multicolumn{8}{c}{\textbf{Object Addition}} \\
\midrule
InsV2V \citep{insv2v}      & 2.71 & 5.31 & 4.84 & 2.04\%  & 25.50 & 0.985 & 0.966 \\
Señorita \citep{senorita-2m}    & 2.63 & 5.43 & 4.80 & 6.12\%  & 25.26 & \underline{0.990} & \underline{0.981} \\
VACE \citep{vace}        & 7.12 & 5.40 & 7.38 & 30.61\% & 28.01 & \textbf{0.990} & 0.980 \\
ICVE \citep{icve}        & \underline{8.95} & \underline{8.65} & \textbf{8.33} & \underline{77.55\%} & \underline{29.13} & 0.987 & 0.974 \\
Lucy Edit \citep{lucyedit}   & 6.96 & 7.29 & 6.78 & 44.90\% & 27.39 & 0.987 & 0.978 \\

\rowcolor[HTML]{fadbd8}
\textbf{VideoCoF (Ours)} & \textbf{9.12} & \textbf{8.78} & \underline{8.27} & \textbf{79.59\%} & \textbf{29.60} & 0.988 & \textbf{0.982} \\
\midrule

\multicolumn{8}{c}{\textbf{Object Swap}} \\
\midrule
InsV2V \citep{insv2v}      & 1.52 & 7.37 & 6.54 & 0.00\%  & 26.22 & 0.991 & 0.984 \\
Señorita \citep{senorita-2m}    & 1.69 & 7.39 & 6.40 & 0.00\%  & 25.97 & \underline{0.994} & 0.990 \\
VACE \citep{vace}        & 8.11 & 6.53 & 7.79 & 34.62\% & \underline{26.93} & \textbf{0.995} & \underline{0.992} \\
ICVE \citep{icve}        & \underline{9.08} & \textbf{8.40} & \textbf{8.57} & \underline{73.08\%} & 26.54 & 0.993 & 0.989 \\
Lucy Edit \citep{lucyedit}   & 6.81 & 7.58 & 7.50 & 44.23\% & 26.46 & 0.992 & 0.988 \\

\rowcolor[HTML]{d6eaf8}
\textbf{VideoCoF (Ours)}  & \textbf{9.10} & \underline{8.39} & \underline{8.14} & \textbf{80.77\%} & \textbf{27.10} & 0.993 & \textbf{0.996} \\
\midrule

\multicolumn{8}{c}{\textbf{Local Style Transfer}} \\
\midrule
InsV2V \citep{insv2v}      & 6.29 & 7.89 & 6.89 & 19.61\% & 26.19 & 0.992 & 0.987 \\
Señorita \citep{senorita-2m}    & 5.60 & 7.69 & 6.33 & 25.49\% & 25.97 & 0.995 & 0.992 \\
VACE \citep{vace}        & 7.18 & 5.53 & 7.65 & 41.18\% & 27.56 & \underline{0.996} & \textbf{0.994} \\
ICVE \citep{icve}        & \underline{7.75} & 7.89 & \textbf{7.98} & \underline{54.90\%} & \underline{27.64} & 0.994 & 0.991 \\
Lucy Edit \citep{lucyedit}   & 5.12 & 7.05 & 6.73 & 27.45\% & 26.71 & 0.993 & \underline{0.992} \\

\rowcolor[HTML]{d0ece7}
\textbf{VideoCoF (Ours)} & \textbf{8.02} & \textbf{8.29} & \underline{7.71} & \textbf{58.82\%} & \textbf{27.80} & \textbf{0.997} & 0.991 \\
\bottomrule
\end{tabular}
}

\vspace{-1.5em}

\label{tab:full_comparison}
\end{table*}




\end{document}